\documentclass[10pt,journal,compsoc]{IEEEtran}

\usepackage[absolute,overlay]{textpos}
\usepackage{pbox}
\usepackage{everypage}

\newcommand{\stamp}[1][© 2022 IEEE. This is the author's version of the article that has been published in IEEE Transactions on Visualization and Computer Graphics. The final version of this record is available at: \href{https://doi.org/10.1109/TVCG.2022.3141040}{\color{blue}10.1109/TVCG.2022.3141040}]{%
\begin{textblock*}{150mm}(37mm,269mm)
\centering%
\small%
\emph{#1}%
\end{textblock*}%
}

\AddEverypageHook{
  \stamp
}

\usepackage{booktabs,ragged2e}

\usepackage[flushleft]{threeparttable}

\usepackage{amssymb}

\usepackage{caption}
\usepackage{booktabs}
\usepackage{multirow}

\usepackage{amsmath}

\ifCLASSOPTIONcompsoc
  \usepackage[nocompress]{cite}
\else
  \usepackage{cite}
\fi
\ifCLASSINFOpdf
   \usepackage[pdftex]{graphicx}
   \graphicspath{{figures/}{pictures/}{photos/}{images/}{./}}
   \DeclareGraphicsExtensions{.pdf,.png,.jpg,.jpeg}
\else
   \usepackage[dvips]{graphicx}
   \DeclareGraphicsExtensions{.eps}
\fi

\usepackage{mathptmx}                  
\usepackage[hidelinks]{hyperref}

\usepackage[dvipsnames]{xcolor} 

\newcommand{\hl}[1]{#1} 

\newcommand{\cit}[1]{``#1''}

\usepackage{csquotes}
\usepackage{cleveref}

\usepackage{scalerel}
\usepackage{tikz}
\usetikzlibrary{svg.path}
\definecolor{orcidlogocol}{HTML}{A6CE39}
\tikzset{
  orcidlogo/.pic={
    \fill[orcidlogocol] svg{M256,128c0,70.7-57.3,128-128,128C57.3,256,0,198.7,0,128C0,57.3,57.3,0,128,0C198.7,0,256,57.3,256,128z};
    \fill[white] svg{M86.3,186.2H70.9V79.1h15.4v48.4V186.2z}
                 svg{M108.9,79.1h41.6c39.6,0,57,28.3,57,53.6c0,27.5-21.5,53.6-56.8,53.6h-41.8V79.1z M124.3,172.4h24.5c34.9,0,42.9-26.5,42.9-39.7c0-21.5-13.7-39.7-43.7-39.7h-23.7V172.4z}
                 svg{M88.7,56.8c0,5.5-4.5,10.1-10.1,10.1c-5.6,0-10.1-4.6-10.1-10.1c0-5.6,4.5-10.1,10.1-10.1C84.2,46.7,88.7,51.3,88.7,56.8z};
  }
}

\newcommand\orcidicon[1]{\href{https://orcid.org/#1}{\mbox{\scalerel*{
\begin{tikzpicture}[yscale=-1,transform shape]
\pic{orcidlogo};
\end{tikzpicture}
}{|}}}}

\begin{document}

%
\title{FeatureEnVi: Visual Analytics for Feature Engineering Using Stepwise Selection and Semi-Automatic Extraction Approaches}
%
%
%
%

\author{Angelos~Chatzimparmpas~\orcidicon{0000-0002-9079-2376},~\IEEEmembership{Student~Member,~IEEE,}
        Rafael~M.~Martins~\orcidicon{0000-0002-2901-935X},~\IEEEmembership{Member,~IEEE~CS,}
        Kostiantyn~Kucher~\orcidicon{0000-0002-1907-7820},~\IEEEmembership{Member,~IEEE~CS,}
        and Andreas~Kerren~\orcidicon{0000-0002-0519-2537},~\IEEEmembership{Senior~Member,~IEEE}
\IEEEcompsocitemizethanks{
  \IEEEcompsocthanksitem Angelos Chatzimparmpas and Rafael M. Martins are with the Department of Computer Science and Media Technology, Linnaeus University, 35195 V{\"a}xj{\"o}, Sweden. \protect\\
  E-mail: \{angelos.chatzimparmpas, rafael.martins\}@lnu.se.
  \IEEEcompsocthanksitem Kostiantyn Kucher and Andreas Kerren are with the Department of Computer Science and Media Technology, Linnaeus University, 35195 V{\"a}xj{\"o}, Sweden and the Department of Science and Technology, Link{\"o}ping University, 60233 Norrk{\"o}ping, Sweden. \protect\\
  E-mail: \{kostiantyn.kucher, andreas.kerren\}@\{lnu, liu\}.se.
}
\thanks{Manuscript received August 29, 2021; revised December 09, 2021.}}

%
%

\markboth{IEEE TRANSACTIONS ON VISUALIZATION AND COMPUTER GRAPHICS,~Vol.~XX, No.~X, APRIL~2022}%
{Chatzimparmpas \MakeLowercase{\textit{et al.}}: FeatureEnVi: \\ Visual Analytics for Feature Engineering Using Stepwise Selection and Semi-Automatic Extraction Approaches}
\IEEEtitleabstractindextext{%
\begin{abstract}
The machine learning (ML) life cycle involves a series of iterative steps, from the effective gathering and preparation of the data---including complex feature engineering processes---to the presentation and improvement of results, with various algorithms to choose from in every step. Feature engineering in particular can be very beneficial for ML, leading to numerous improvements such as boosting the predictive results, decreasing computational times, reducing excessive noise, and increasing the transparency behind the decisions taken during the training. Despite that, while several visual analytics tools exist to monitor and control the different stages of the ML life cycle (especially those related to data and algorithms), feature engineering support remains inadequate. 
In this paper, we present FeatureEnVi, a visual analytics system specifically designed to assist with the feature engineering process. Our proposed system helps users to choose the most important feature, to transform the original features into powerful alternatives, and to experiment with different feature generation combinations. Additionally, data space slicing allows users to explore the impact of features on both local and global scales. FeatureEnVi utilizes multiple automatic feature selection techniques; furthermore, it visually guides users with statistical evidence about the influence of each feature (or subsets of features). The final outcome is the extraction of heavily engineered features, evaluated by multiple validation metrics.
%
The usefulness and applicability of FeatureEnVi are demonstrated with two use cases and a case study.
We also report feedback from interviews with two ML experts and a visualization researcher who assessed the effectiveness of our system.
\end{abstract}

\begin{IEEEkeywords}
Feature selection, feature extraction, feature engineering, machine learning, visual analytics, visualization
\end{IEEEkeywords}}

\stamp

\maketitle

\IEEEdisplaynontitleabstractindextext

%
\IEEEpeerreviewmaketitle

\IEEEraisesectionheading{\section{Introduction} \label{sec:intro}}%
  \begin{figure*}[ht]
 \centering
 \includegraphics[width=\linewidth]{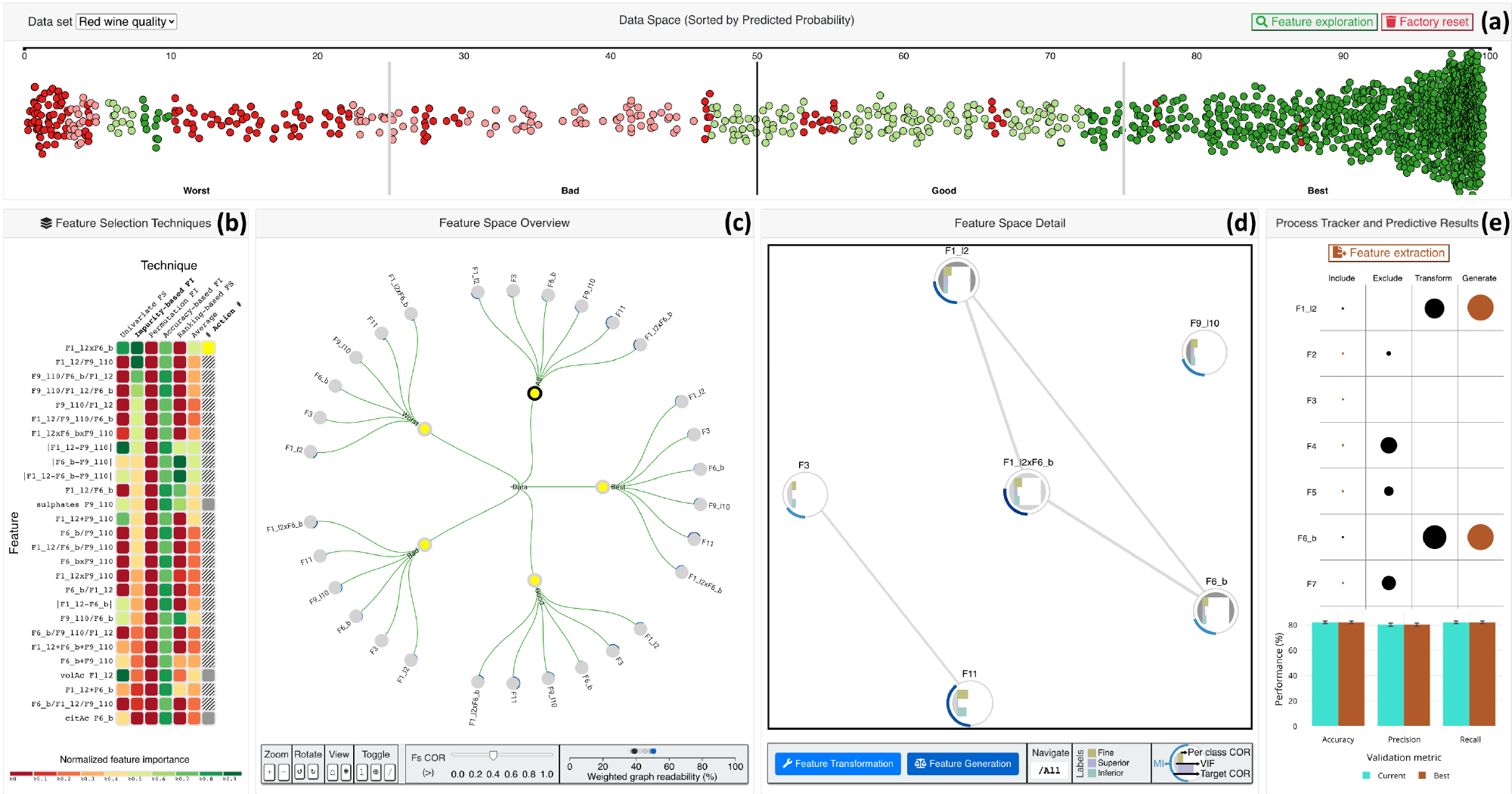} \vspace{-5mm}
 \caption{Selecting important features, transforming them, and generating new features with FeatureEnVi: (a) the horizontal beeswarm plot for manually slicing the data space (which is sorted by predicted probabilities) and continuously checking the migration of data instances throughout the process; (b) the table heatmap view for the selection of features according to feature importances calculated from automatic techniques; (c) the radial tree providing an overview of the features with statistical measures for the different groups of instances, as set by the user-defined data slices; (d) the graph visualization for the detailed exploration of features, their transformation, and comparison between two or three features for feature generation purposes; and (e) the punchcard for tracking the steps of the process and the grouped bar chart for comparing the current vs. the best predictive performance based on three validation metrics.}
 \label{fig:teaser}
\end{figure*}

\IEEEPARstart{I}n machine learning (ML), \emph{classification} is a type of supervised learning where the primary goal is to predict the dependent variable---also known as the target or class label---of every data instance (e.g., rows in a table) given independent features of the data (e.g., columns in a table). \textbf{Feature engineering} is the process of converting raw data into a set of features that better expresses the underlying problem, resulting in powerful ML models with enhanced predictive performance based on validation metrics~\cite{Khurana2018Feature}. In practice, most of the time spent in ML is in preparing this set of features, which should be as concise as possible while retaining vital information about the data set~\cite{Domingos2012A,Kandel2012Enterprise}. If complications occur in this step and remain undetected, they can spoil the later phases of the ML pipeline, according to the classic ``garbage in, garbage out'' principle.  Another important reason why feature engineering is essential in real-world problems is that it increases the transparency and trustworthiness of the data and, in consequence, the ML process in general~\cite{Hong2020Human}. This matter has been drawing attention recently with, for example, the new European General Data Protection Regulation (GDPR) instructions~\cite{GDPR}.
Furthermore, domain experts are increasingly requesting clear evidence in order to trust in ML~\cite{Zhou20182D}. 

In general, feature engineering can be subdivided into four major processes: (a) \emph{feature ideation}, (b) \emph{feature generation}, (c) \emph{feature transformation}, and (d) \emph{feature selection}~\cite{Heaton2016An,Brooks2015FeatureInsight}. Feature ideation is the process of coming up with entirely new features from the \cit{raw} data. It is heavily subjective for most applications, except for text data~\cite{Brooks2015FeatureInsight,Cheng2015Flock}, for instance. As we work with numerical values and tabular data, the focus of our analytical approach presented in this paper is on the last three categories, which we describe next.

Feature generation is the action of creating more useful features from the combination of already existing ones. Typical interactions during this process are associated with the experimentation of: (1) \emph{addition}, (2) \emph{subtraction}, (3) \emph{multiplication}, and (4) \emph{division} between correlated features (all these operations are supported by our approach)~\cite{Huan1998Feature}. ML experts and practitioners usually mix various features in a trial-and-error manner until the new features meet their expectations~\cite{Patel2008Investigating,Domingos2012A}. The alternative is to use automatic feature generation techniques, but they are computationally expensive and may take a long time to produce an outcome~\cite{Patel2008Investigating,Markovitch2002Feature,Schuller2006Evolutionary}. Another threat in the validity of automatic feature generation is that the optimization is performed according to a single measurement, such as information gain, for example~\cite{Domingos2012A}. Additionally, features that look irrelevant in isolation may be informative in combination. Thus, one question that remains open is: \textbf{(RQ1)} for a given data set, which features should we compare, and how should we combine them to generate a new feature that boosts performance?

Feature transformation usually denotes less sophisticated modifications over the features~\cite{Kankanige2014Improved}. Some of the standard transformations also supported by our approach are: (1) \emph{rounding}, (2) \emph{binning}, (3) \emph{scaling}, (4) \emph{logarithmic transformations}, (5) \emph{exponential transformations}, and (6) \emph{power functions}. In this scenario, ML experts and practitioners usually perform exploratory data analysis (EDA), sometimes utilizing visualization of the data distribution to understand which transformation they should apply~\cite{Muhlbacher2013APartition}. Another option is to use automated feature transformation approaches such as a transformation graph~\cite{Khurana2018Feature}, which can, however, result in overfitting~\cite{Domingos2012A}. Unfortunately, such methods were only employed in regression and reinforcement learning problems~\cite{Khurana2018Feature}. This directs us to an additional open question: \textbf{(RQ2)} which features should we transform, and how can we understand their impact on the final outcome when using a specific data set? 

Feature selection is about choosing a subset of features from the pool of features available by that time. Feature selection methods can be generally divided into four high-level categories: (1) \emph{filter methods}, (2) \emph{wrapper methods}, (3) \emph{embedded methods}, and (4) \emph{hybrid methods}~\cite{Guyon2003An,Chandrashekar2014A,Kohavi1997Wrappers}. Our feature selection strategy belongs to the last category, as we incorporate techniques from all the other categories. Also, instead of appending features progressively (called \emph{forward selection}) or considering all features and then discarding some (known as \emph{backward elimination}), we choose a \emph{stepwise selection approach}. Therefore, we start with all features, but we can add or remove any number of features at different stages. For feature selection, the vast majority of the existing literature is on automatic feature selection techniques. They may, however, lack transparency without the assistance of visualizations~\cite{Blum1997Selection,Guyon2003An,Chandrashekar2014A,Duan2005Multiple,Ding2011Svm}. Furthermore, there is an opportunity to select features from a candidate set, which can be time-consuming if this set is large~\cite{Yang1997A,Aphinyanaphongs2014A,Forman2003An}. Even though a series of analytical tools and systems have been developed to address such challenges with the use of visualizations~\cite{May2011SmartStripes,May2011Guiding,Dennig2019FDive}, a remaining open question is: \textbf{(RQ3)} which feature selection technique should we follow when they present diverging results, and how can we verify their effectiveness for particular problems?


The complex nature of feature engineering, occasionally declared as ``black art'' \cite{Domingos2012A,Duboue2020The}, motivated us to concentrate our effort on addressing the three research questions mentioned above. In this paper, we present a visual analytics (VA) system, called \emph{FeatureEnVi} (\textbf{Feature} \textbf{En}gineering \textbf{Vi}sualization, as seen in \autoref{fig:teaser}), that utilizes stepwise selection and semi-automatic feature extraction approaches for the feature engineering process of a state-of-the-art ensemble learning algorithm known as \emph{XGBoost} \cite{Chen2016XGBoost}. FeatureEnVi uses validation metrics for multi-class classification problems, and includes three core iterative feature engineering phases that are intertwined and monitored by automatic techniques and statistical measures (see the details in \autoref{sec:system}).
To provide answers to the three research questions (\textbf{RQ1--RQ3}), FeatureEnVi adopts the following \textbf{workflow} (cf. \autoref{fig:workflow-diagram} present in \autoref{sec:system}): (i) the initialization of the feature exploration after the user has split the data into four distinct slices sorted by predicted probabilities, (ii) the feature selection subprocess with five automated techniques computed for the choice of features, (iii) the feature transformation that enables the user to convert features to impactful alternatives while utilizing statistical measures, (iv) the feature generation for the creation of new features from the combination of already existing features, and (v) the tracking of feature changes and analysis of the previous phases using visualizations, while the performance of the current solution is compared to the best result achieved so far.
Overall, our contributions consist of the following:

\begin{itemize}
\item the definition of a unified visual analytic workflow for feature engineering, fusing stepwise selection and semi-automatic extraction approaches;
\item a prototype implementation of the proposed workflow, in the form of our VA system, called FeatureEnVi, that consists of a novel combination of multiple coordinated views to support the complete process of engineering a data set's features;
\item a proof-of-concept of the applicability of our proposed system with two use cases and a case study, using real-world data, that verifies the usefulness of our choice to deploy multiple statistical measures and include the human in-between automated methods; and
\item the discussion of the methodology and results of interview sessions with ML experts and a visualization researcher, presenting encouraging results.
\end{itemize}          

\noindent The rest of this paper is organized as follows. In~\autoref{sec:relwork}, we review automatic feature generation methods, then we continue with feature transformations, and finally, automated and visually-assisted selection of subsets of features. Afterwards, in~\autoref{sec:goals}, we describe the analytical tasks and design goals for applying VA for feature engineering, and we highlight the necessity for multiple statistical measures. 
\autoref{sec:system} presents the system's functionalities and simultaneously describes the first use case to remove unnecessary features, transform influential features to improve the performance, and generate new features for a further increase of the predictive score using physicochemical data
; it also includes the second case, in which we demonstrate the applicability and usefulness of FeatureEnVi with a real-world data set focusing on vehicle recognition. 
In~\autoref{sec:case}, we describe a case study carried out with an ML expert, focusing on improving classification for the molecules biodegradation problem.
Next, in~\autoref{sec:eval}, we analyze the feedback on our VA system acquired from the expert interview sessions, including a discussion of limitations that lead us to potential future directions for FeatureEnVi.
Subsequently, in~\autoref{sec:disc}, we discuss alternative design choices for our visual representations as well as additional limitations and opportunities for future work. 
Finally,~\autoref{sec:con} concludes our paper.


\section{Related Work} \label{sec:relwork}%
  Several VA systems have been developed to explore and select subsets of features with the help of visualization. Finding which features to transform, and how, together with generating new features from different combinations, are some of the core phases that lack attention from the InfoVis/VA communities.
This section reviews prior work on feature generation, feature transformation, and feature selection techniques for both automated and visually-assisted feature engineering. We also discuss the differences of such tools compared to FeatureEnVi to emphasize our system's novelty. To the best of our knowledge, there is no literature explaining the use of VA for the complete feature engineering process (as specified in~\autoref{sec:intro}) along with the partitioning of the data space based on the extracted predicted probabilities. 

Additionally, a summary of the tools discussed in this section and comparison of their capabilities with FeatureEnVi are presented in~\autoref{capcomp}. A tick indicates that the tool has the corresponding capabilities, while a tick in parentheses means the tool offers implicit support (i.e., it could be done manually, in an ad hoc manner, but is not explicitly supported). The table does not include works that do not contain a concrete visualization tool as their research contribution, as in the work by Khurana et al.~\cite{Khurana2018Feature}, for instance. This table thus highlights the similarities/dissimilarities of FeatureEnVi against previously developed VA tools. From \autoref{capcomp}, it is evident that no other VA tool offers all functionalities of our system and is directly comparable to FeatureEnVi.

\begin{table}[tb]
\captionsetup{justification=centering, labelsep=newline}
\begin{threeparttable}
\caption{Capabilities Comparison of FeatureEnVi With Other Tools}
\label{capcomp}
\setlength\tabcolsep{0pt} 

\begin{tabular*}{\columnwidth}{@{\extracolsep{\fill}} l cccc}
\toprule
     Tool &
     \multicolumn{4}{c}{Feature Engineering} \\
\cmidrule{2-5}
     & Ideation & Generation & Transformation & Selection \\
\midrule
     FeatureEnVi & & \checkmark & \checkmark & \checkmark \\
     M{\"u}hlbacher and Piringer~\cite{Muhlbacher2013APartition} & & & \checkmark & \checkmark \\
	 FeatureInsight~\cite{Brooks2015FeatureInsight} & \checkmark & & & (\checkmark) \\
	 Flock~\cite{Cheng2015Flock} & \checkmark & & & (\checkmark) \\
	 Liu et al.~\cite{Liu2018Visual} & & (\checkmark) & & (\checkmark) \\
	 \emph{Other works in~\autoref{sec:select}} & & & & \checkmark \\
\bottomrule
\end{tabular*}
\smallskip
\scriptsize
\footnotesize{The tick in parentheses (\checkmark) indicates partial support of a capability.}\\
\end{threeparttable}
\end{table}

\subsection{Feature Generation} \label{sec:generate}

As mentioned in~\autoref{sec:intro}, many studies confirm the importance of high-quality features in ML (e.g., the works by Domingos~\cite{Domingos2012A} or Yang and Pedersen~\cite{Yang1997A}). Practitioners often spend a substantial amount of time experimenting with custom combinations of features or utilizing algorithmic feature generation~\cite{Patel2008Investigating,Domingos2012A}. In several application domains, algorithms have been useful for creating high-level features from data elements related to speech~\cite{Schuller2006Evolutionary}, real-time sensors~\cite{Fogarty2007Toolkit}, and general ML problems~\cite{Markovitch2002Feature}. However, there is no single metric that is consistently suitable across all data sets. 
Therefore, certain levels of involvement and manual effort are still required from analysts, and they receive little to no guidance during this complicated procedure.

A use case present in a visual diagnosis tool revealed that feature generation involving the combination of two features is capable of a slight increase in performance~\cite{Liu2018Visual}. The authors tested the same mathematical operations as in our system (i.e., addition, subtraction, multiplication, and division), but the generation was performed manually by the analysts. Also, the decision for this action was based solely upon the similarity in those features' distributions~\cite{Liu2018Visual}. 
In FeatureEnVi, determining which features to match during feature generation is achieved by analyzing linear and nonlinear relations present in the data. For the former, one of the most well-known approaches is \emph{Pearson's correlation coefficient} between features and with the target variable~\cite{Zhao2019FeatureExplorer,Rojo2020GaCoVi}. For the latter, \emph{mutual information} is used in our VA system (also used by May et al.~\cite{May2011Guiding}, for instance). Features are added to capture the missing information and improve the classifier's performance~\cite{Guo2003Coordinating}. The magnitude of correlation with the dependent variable and in-between features is key to such decisions~\cite{Perez2017Accurate,Hall2000Correlation}. However, the aforementioned VA tools work with regression problems and only support feature selection.

\subsection{Feature Transformation} \label{sec:transform}

Automatic feature transformation has been examined within the ML community with positive results in reinforcement learning. In the work by Khurana et al.~\cite{Khurana2018Feature}, the authors conduct a performance-driven exploration of a transformation graph which systematically enumerates the space of given options. A single ``best'' measurement is not possible, however, since the options might conflict with each other. In contrast, FeatureEnVi focuses on classification problems and presents users with various statistics about each feature in four different slices of the data space (the ones considered in~\autoref{sec:generate}, along with \emph{variance influence factor} and \emph{per class correlation}). This is similar to ExplainExplore\cite{Collaris2020ExplainExplore} for classification and HyperMoVal~\cite{Piringer2010HyperMoVal} for regression. The authors of these tools work with the probabilities of instances to belong in different classes or clusters depending on the features. In our case, we take into account the ground truth values of the target variable and compute the probability (0\% to 100\%) of the ML model to classify each instance correctly.

A VA system for regression analysis has been proposed by M{\"u}hlbacher and Piringer~\cite{Muhlbacher2013APartition}. The system is more similar to our work as it also incorporates feature transformation in its design (specifically logarithmic, exponential, and power functions). The main difference between this work and ours is our focus on classification rather than regression models. Moreover, they compared only the distribution of data values per feature to choose which transformation is better. In FeatureEnVi, we use multiple statistical measures to understand the relation of features with their transformed counterparts.

\subsection{Feature Selection} \label{sec:select}

There is a rather large body of existing work on automatic feature selection techniques~\cite{Guyon2003An,Blum1997Selection,Chandrashekar2014A}. However, one limitation is that features can be redundant if there is a strong correlation among them, and the correlation coefficient is unable to characterize nonlinear relationships. Thus, this is a problem where the feature selection techniques struggle to find a solution because of multiple parameters they have to optimize simultaneously. Guyon and Elisseeff~\cite{Guyon2003An} performed a survey including an extensive description of automatic feature selection pitfalls. The authors stress the general problem of finding the smallest possible subset of features for a given data set. They suggest that an automated method cannot be expected to find the best feature subset in all cases by itself. Other methods that face the same challenge are wrappers that use regression or classification models to find an ideal feature subset by iteratively including or excluding features. The combination of learning models (e.g., SVM~\cite{Hearst1998Support}) and wrapper methods (e.g., RFE~\cite{Guyon2002Gene}) is a commonly used approach for automatic feature selection~\cite{Duan2005Multiple,Ding2011Svm}. Also, metric-based ranking followed by the selection of the $k$ best features~\cite{Aphinyanaphongs2014A,Forman2003An} and more complex metrics---such as those used in genetic algorithms~\cite{Yang1998Feature}---have been examined in the past. But they suffer from the same issues as described before. In our VA system, we implement several alternative feature selection techniques belonging to different types, and we allow users to decide if their aggregation is ideal or they want to focus on one of them. 

Various visualization techniques have been proposed for the task of feature selection, including correlation matrices~\cite{Friendly2002Corrgrams,MacEachren2003Exploring}, radial visualizations~\cite{Sanchez2018Scaled,Artur2019A,Wang2017Linear}, scatterplots~\cite{Yang2004Value}, scatterplot matrices~\cite{Elmqvist2008Rolling}, feature ranking~\cite{Yang2003Visual,Krause2014INFUSE,Piringer2008Quantifying,Seo2005A,Turkay2011Brushing,Johansson2009Interactive,Lu2014Integrating,Gratzl2013LineUp}, feature clustering~\cite{Yang2003Interactive}, and dimensionality reduction (DR)~\cite{Turkay2011Brushing,Lin2018RCLens,Rauber2015Interactive}. The category of techniques more related to our work is \emph{feature ranking}, since we use automatic feature selection techniques to rank the importance of the different features. For example, a VA tool called INFUSE~\cite{Krause2014INFUSE} was designed to aid users in understanding how features are being ranked by the automated feature selection techniques. It presents an aggregated view of results produced by automatic techniques, assisting the user in learning how these work and compare their results with multiple algorithms. Similarly, Klemm et al.~\cite{Klemm20163D} propose an approach that performs regression analysis exhaustively between independent features and the target class. These approaches take into account the user’s ability to identify patterns from analyzing the data (e.g., with the colors in a heatmap representation) or choose the feature subset by some quantitative metric. A few other VA systems have leveraged a balanced blending between automatic and visual feature selection techniques. RegressionExplorer~\cite{Dingen2019RegressionExplorer} is one example for examining logistic regression models. Additionally, the exploration of linear relationships among features was studied by Barlowe et al.~\cite{Barlowe2008Multivariate}. FeatureEnVi offers rather similar characteristics with the tools analyzed above. However, we combine several automatic feature selection techniques and statistical heuristics in cohesive visualizations for evaluating feature selection and feature extraction concurrently.

Visual support for the task of feature subset selection requires displaying information on different levels of granularity; highly detailed views are not optimal because they do not scale well with many features. For instance, the tool by Hohman et al.~\cite{Hohman2020Understanding} facilitates the visual comparison of feature distributions for a high number of features. It visually exposes the divergence of distributions in terms of training and testing splits, predictive performance (correct vs. incorrect predictions), and multiple data versions. Overall, this tool is less related to the context of our work because it focuses on the challenge of data iteration, in which analysts continuously add and/or remove instances over iterations. Bernard et al.~\cite{Bernard2014Visual} present a tool that forms a connection panorama between features and between their bins. However, their VA tool is limited to categorical data. Another feature selection approach is SmartStripes by May et al.~\cite{May2011SmartStripes}. It supports the investigation of interdependencies between different features and entity subsets. Their main idea is to find features with the strongest correlation with instance partitions of the data set. Alternatively, our approach groups instances based on the predicted probability of an ML algorithm. It also indicates features that played a vital role in distinguishing instances from the worst to the best predicted.


\section{Analytical Tasks and Design Goals} \label{sec:goals}
  This section specifies the fundamental analytical tasks (\textbf{T1--T5}) that users should be capable of when performing feature engineering while they obtain assistance and guidance from a VA system. Afterwards, we report the corresponding design goals (\textbf{G1--G5}) that led to the development of our proposed FeatureEnVi system. 

\subsection{Analytical Tasks for Feature Engineering}

We derived the analytical tasks described in this section from the in-depth analysis of the related work in~\autoref{sec:relwork}. The three analytical tasks from Krause et al.~\cite{Krause2014INFUSE}, the three experts who expressed their requirements in Zhao et al.~\cite{Zhao2019FeatureExplorer}, and the user tasks acquired through expert interviews from Collaris and van Wijk~\cite{Collaris2020ExplainExplore} also inspired our work. Additionally, our own recently conducted surveys~\cite{Chatzimparmpas2020A,Chatzimparmpas2020The} and experiences with prototyping VA tools for ML~\cite{Chatzimparmpas2020t,Chatzimparmpas2021StackGenVis,Chatzimparmpas2021Visevol} played a significant role in the final formulation of the following analytical tasks.

\textbf{T1: Inspect the impact of features in both local and global perspectives.} 
In data sets with many features, it is not trivial to acknowledge each feature's contribution in the final prediction. Notably, a user needs a global mechanism to examine the relationships between those features (if there are any). Before users decide about the engineering strategy behind every feature, another factor that they must consider is the impact of features in representative groups of instances~\cite{Collaris2020ExplainExplore}. Finding such groups should be a goal for VA systems (cf. \textbf{G1}).

\textbf{T2: Contrast various feature selection techniques.}
The results produced by automatic feature selection techniques can be remarkably inconsistent. A user should be able to compare alternative feature selection techniques according to their rankings for all features. The interviewees of Krause et al.~\cite{Krause2014INFUSE} stated that they would like to get answers to questions such as: ``Which features are consistently ranked highly?'', ``How much do the techniques differ in their ranking?'', and ``Are there features that have a high rank with some techniques and a low rank with some others?'' The visual feedback from VA systems should provide users the ability to respond to those inquiries (\textbf{G2}).

\textbf{T3: Analyze the effect of diverse feature transformations.}
When some features' distributions are skewed left or right, logarithmic or exponential transformations respectively could further boost each feature's contribution. Moreover, particular ML algorithms (e.g., SVM~\cite{Hearst1998Support} and XGBoost~\cite{Chen2016XGBoost}) might expect a few preprocessing steps, for instance, scaling the data values with $z$-score normalization~\cite{Altman1968Financial}. A user should be competent in judging the influence of feature transformations before applying them~\cite{Collaris2020ExplainExplore}. As explained in \textbf{G3}, visual clues are mandatory for assessing every transformation in practice.

\textbf{T4: Combine features to generate new features.} 
It is common in feature engineering to combine features in many ways by using mathematical operators. Feature generation produces features with higher levels of contribution because they pack additional information. Here, users need means to select a few features and a way to evaluate the new combination. A typical question connected to \textbf{T2} (about feature subset selection) and this task was made by the participants of Krause et al.'s interview session~\cite{Krause2014INFUSE}: “How does the performance of the model increase or decrease if I remove or add these features?” The same could apply for newly-generated features (that we are unaware of) when compared against the original pool of features (see \textbf{G4}).

\textbf{T5: Evaluate the results of the feature engineering process.} 
At any stage of the feature engineering process (\textbf{T2--T4}), a user should be able to observe the fluctuations in performance with the use of standard validation metrics (e.g., \emph{accuracy}, \emph{precision}, and \emph{recall})~\cite{Zhao2019FeatureExplorer}. Also, users could possibly want to refer to the history of the actions they performed to identify the key spots that might have corresponded to improvements in the outcome. VA systems must thus be able to provide ways of monitoring the performance, as described in \textbf{G5}.

\subsection{Design Goals for FeatureEnVi}

To satisfy the previously defined analytical tasks, we have determined five design goals to be addressed by our tool/system. The implementation of these design goals is illustrated in \autoref{sec:system}.

\textbf{G1: Division of data space into slices based on predicted probabilities, for transparent local feature contribution.} 
Our goal is to assist in the search for distinctive features that might contribute more to instances that are harder or easier to classify~(\textbf{T1}). By splitting the data space into quadrants, we aim to assure that users' interference with the engineering of specific features does not cause problems in key parts of the data space. The tool should begin by training an initial model with the original features of each data set, which will be a starting point (i.e., ``state zero'') for users to compare future interactions and take further decisions.

\textbf{G2: Deployment of different types of feature selection techniques to support stepwise selection.}
There are several different techniques for computing feature importance that produce diverse outcomes per feature. The tool should facilitate the visual comparison of alternative feature selection techniques for each feature (\textbf{T2}). Another key point is that users should have the ability to include and exclude features during the entire exploration phase.

\textbf{G3: Application of alternative feature transformations according to feedback received from statistical measures.}
In continuation of the preceding goal, the tool should provide sufficient visual guidance to users to choose between diverse feature transformations (\textbf{T3}). Statistical measures such as \emph{target correlation} and \emph{mutual information} shared between features, along with \emph{per class correlation}, are necessary to evaluate the features' influences in the result. Also, the tool should use \emph{variance influence factor} and \emph{in-between features' correlation} for identifying colinearity issues. When checking how to modify features, users should be able to estimate the impact of such transformations.

\textbf{G4: Generation of new features and comparison with the original features.}
With the same statistical evidence as defined in \textbf{G3}, users should get visually informed about strongly correlated features that perform the same for each class. Next, the tool can use automatic feature selection techniques to compare the new features with the original ones, using the same methods as in \textbf{G2}. Finally, the tool should let users select the proper mathematical operation according to their prior experience and the visual feedback (\textbf{T4}).

\textbf{G5: Reassessment of the instances' predicted probabilities and performance, computed with appropriate validation metrics.} In the end, users' interactions should be tracked in order to preserve a history of modifications in the features, and the performance should be monitored with validation metrics (\textbf{T5}). At all stages, the tool should highlight the movement of the instances from one data slice to the other due to the new prediction probability values of every instance. The best case is for all instances to relocate from the left to the right-most side.

\section{FeatureEnVi: System Overview and Application} \label{sec:system}%
  \begin{figure}[tb]
\centering
\includegraphics[width=\linewidth]{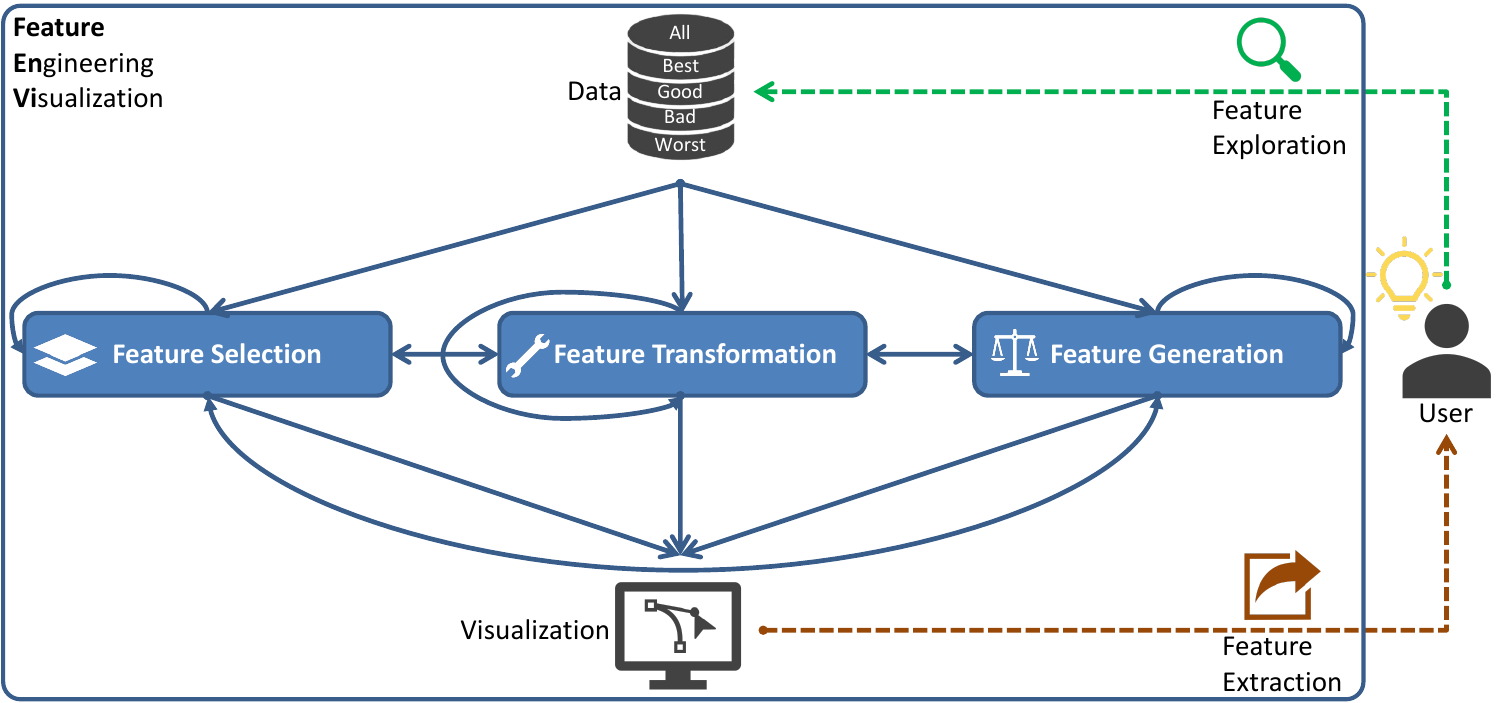} \vspace{-5mm}
\caption{The FeatureEnVi workflow begins with partitioning the data set according to the prediction probabilities of instances. The data is passed to three different feature engineering processes (selection, transformation, and generation) which are executed iteratively, under the control of the user through the interface.
}
\label{fig:workflow-diagram}
\end{figure}

\begin{figure*}[tb]
\centering
\includegraphics[width=\linewidth]{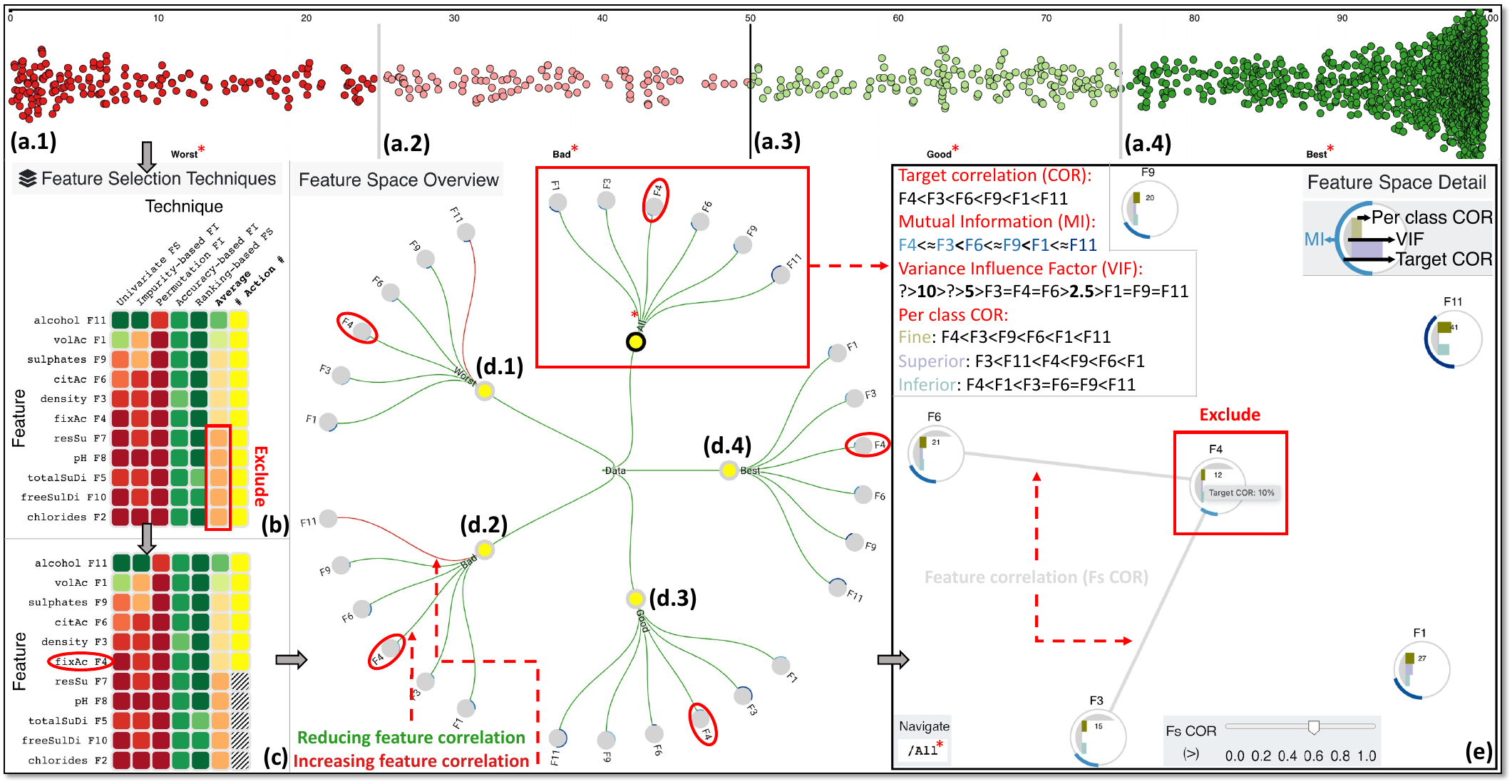} \vspace{-5mm}
\caption{Exploration of features with FeatureEnVi. The \emph{default} slicing thresholds for the data space separate the instances into four quadrants that represent intervals of 25\% predicted probability (see (a.1--a.4)). View (b) presents a table heatmap with five different feature selection techniques and their average value per feature. We exclude the less contributing features, as shown in the duplicated view (c). In the radial tree, the paths from (d.1) to (d.4) are the features for the groups formed at (a.1--a.4), respectively, while the features' impact for the entire data set is shown in the red box. The whole data space is displayed with even more details in the graph visualization in (e), where additional metrics' results are reported. A summary of the meaning of the visual encodings for these metrics is visible in the top-left corner in (e). More details about these views are described in the text.}
\label{fig:use_case1_sel}
\end{figure*}

\begin{figure*}[tb]
  \centering
  \includegraphics[width=\linewidth]{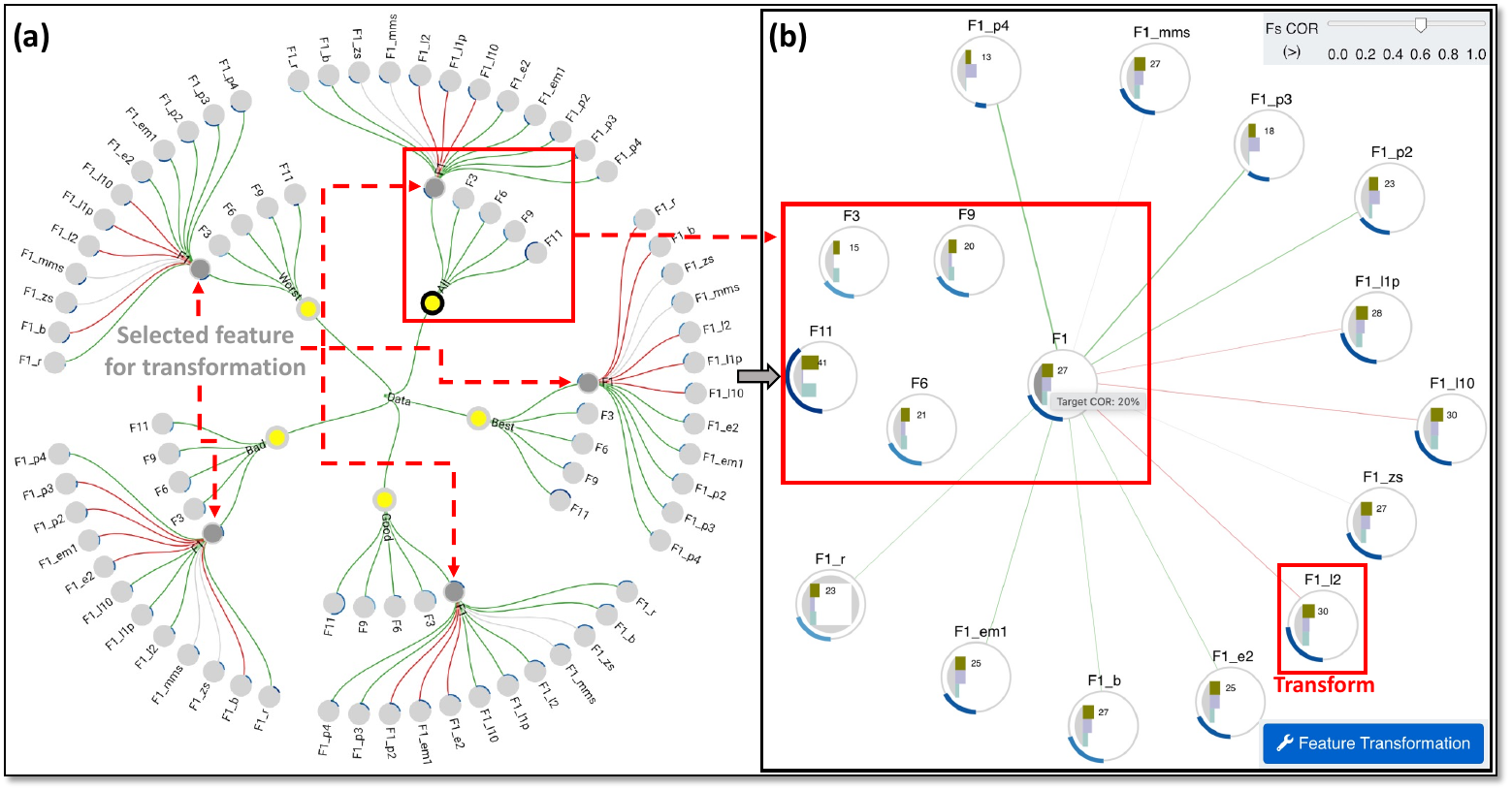} \vspace{-5mm}
  \caption{The feature transformation phase in order to empower features and improve the prediction. In (a), we have selected F1, and we check the impact of different transformation strategies in all slices of the data space. The entire data space is shown in more detail with the graph visualization in (b). The F1's feature transformation that appears the best is the logarithmic transformation (\emph{\_l2} or \emph{\_l10}).}
  \label{fig:use_case1_tra}
\end{figure*}

Following the analytical tasks and our derived design goals, we have developed FeatureEnVi, an interactive web-based VA system that allows users to utilize multiple metrics and automatic methods in order to engineer useful features.

The system consists of five main interactive visualization panels (\autoref{fig:teaser}): (a) data space ($\rightarrow$  \textbf{G1}), (b) feature selection techniques ($\rightarrow$  \textbf{G2}), (c) feature space overview, (d) feature space detail ($\rightarrow$  \textbf{G3} and \textbf{G4}), and (e) process tracker and predictive results ($\rightarrow$  \textbf{G5}). 
We propose the following \textbf{workflow} for the integrated use of these panels (cf. \autoref{fig:workflow-diagram}): 
(i) choose four suitable data space slices, which are then used for evaluating the impact of each feature on particular groups of instances (\autoref{fig:teaser}(a));
(ii) in the exploration phase, choose subsets of features using diverse automatic feature selection techniques (see \autoref{fig:teaser}(b));
(iii) in the overview phase, explore the impact of features in both the individual slices and the entire data space, with various computed metrics (\autoref{fig:teaser}(c));
(iv) during the detailed examination phase, check the different transformations of the features with statistical measures and compare the combinations of two or three features that result in newly-generated features (cf. \autoref{fig:teaser}(d)); and
(v) contrast the performances of the \emph{best} predictive performance found so far vs. the \emph{current} result according to three validation metrics in \autoref{fig:teaser}(e).
All these processes are iterative and could happen in any other order. The final outcome is the generated knowledge acquired from the extracted features. Note that the typical workflow, which is followed in the two use cases and the case study in~\autoref{sec:case}, 
is linear to the design of the views of FeatureEnVi (i.e., from \textbf{G1} to \textbf{G5}). The individual panels and the workflow are discussed in more detail below.

All visual encodings designed for the panels of FeatureEnVi are summarized in~\autoref{tab:encode}. On the right-hand side, we can observe the optimal states for the available statistical measures. However, in reality, many of the statistical measures will be contradictory to each other, and human decisions are essential on such occasions. 
\hl{To the best of our knowledge, little empirical evidence exists for choosing a particular measure over others. In general, target correlation and mutual information (both related to the influence between features and the dependent variable) may be good candidates for identifying important features~\cite{Li2004Efficient}. After these first indicators, the remaining collinearity measures can be useful too. Although it can be claimed that a high level of correlation between features is from 0.9 and above~\cite{Dohoo1997An}, no precise rules exist for judging the significance of the collinearity (see our discussion on the variance influence factor in~\autoref{sec:featdet}). To accomodate for these challenges, a variety of measures are simultaneously visualized in our tool.} 
In detail, for the \emph{Data Space} panel, the predicted probabilities should become dark or light green. The more important a feature, it is more likely to impact the outcome of the ML model, thus, in the \emph{Feature Selection Techniques} panel, the normalized importance should be mostly green or close to green colors. The \emph{Feature Space Overview} panel contains a subset of the measures present in the \emph{Feature Space Detail} panel. For the former, target correlation \emph{(COR)} should be---as high as possible---depicted with a full circle bar. The same applies to mutual information \emph{(MI)}, but the indication here is the dark or darker blue color. For the difference between features correlation, the optimal is to reduce, hence, green is the expected outcome. The same encodings also apply for the latter panel with the addition of the variance influence factor \emph{(VIF)}, per class correlation \emph{(COR)}, and between features correlation \emph{(Fs)}. The per class correlation is the only of the three measurements that should be substantially high, which is illustrated with long bars for the horizontal bar chart. The remaining should both be decreasing, as a result, the gray color visible in this panel should be diminishing---as much as possible. Finally for the \emph{Process Tracker and Predictive Results} panel, moving forward to the next exploration step is mapped to bigger in size circles with brown color used for the best setup of features explored so far.

\begin{table}[tb]
  \captionsetup{justification=centering, labelsep=newline}
  \centering
  \caption{Summary of the Various Measurements and their Visual Mappings for the Different Panels Supported by FeatureEnVi} \vspace{-1mm}
  \label{tab:encode}
  \includegraphics[width=\columnwidth]{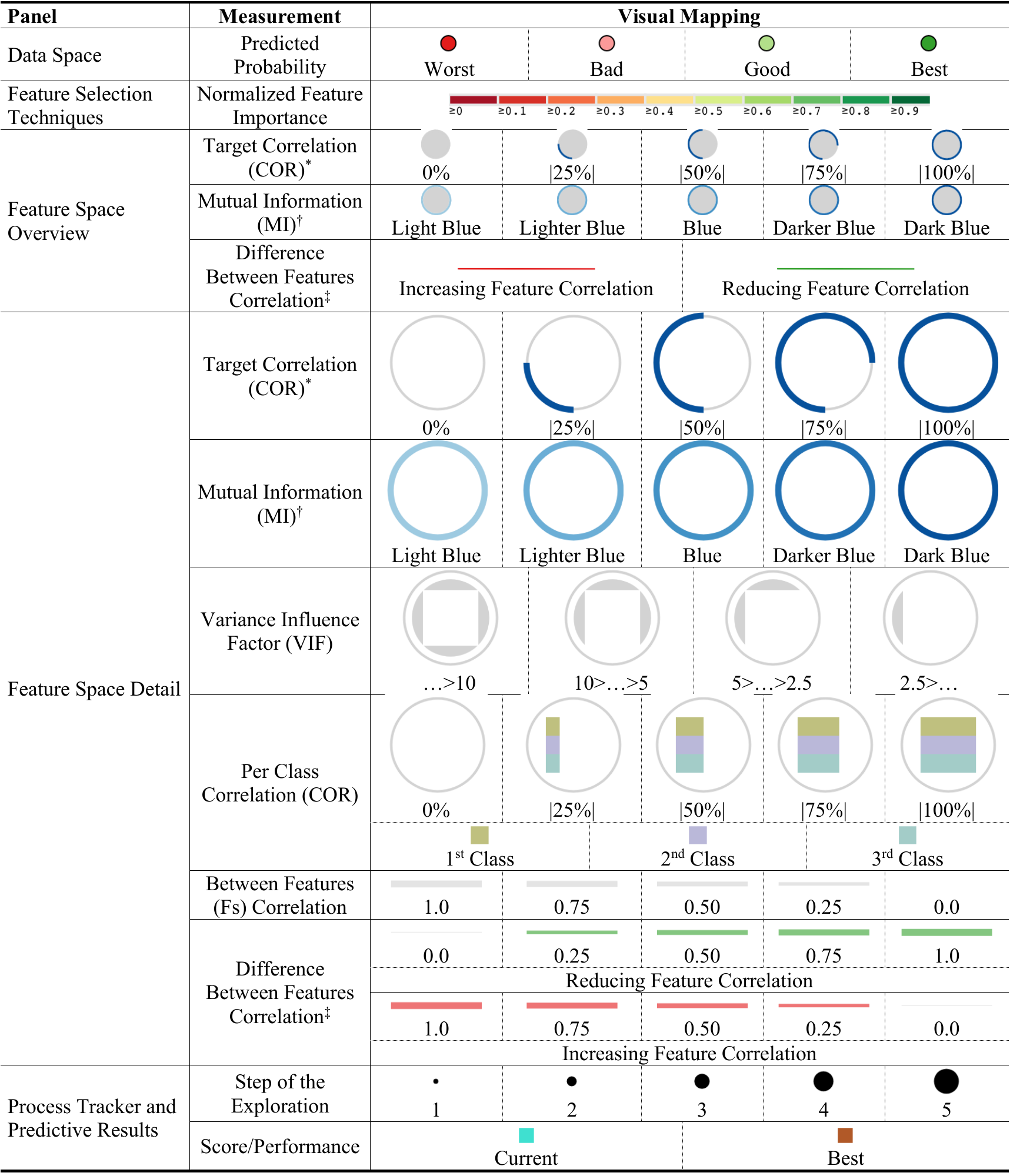} \vspace{-3mm}
  \begin{tablenotes}[para,flushleft]
  \footnotesize{This table is hierarchically organized according to the five main panels that follow our system’s workflow. From left to right, we present the worst to the best possible outcome for all measurements.
  \newline
  $^{*}$, $^{\dagger}$, and $^{\ddagger}$: These measurements are shared across the Feature Space Overview and the Feature Space Detail panels, using similar visual mappings.}\\
  \end{tablenotes} \vspace{-2mm}
\end{table}

The workflow of FeatureEnVi is model-agnostic \hl{within the scope of classification problems}, meaning that it could be paired with any \hl{classification} algorithm.
In this current version, the implementation uses a state-of-the-art ensemble learning method called XGBoost~\cite{Chen2016XGBoost}. This choice was made intentionally because some algorithms (e.g., SVM~\cite{Hearst1998Support} and XGBoost~\cite{Chen2016XGBoost}) are susceptible to specific types of transformations (e.g., scaling). To make our approach even more future-proof, we train this ML algorithm with the \emph{Bayesian Optimization} package~\cite{Bayesian}. For this section, 
we validate our results with cross-validation using 8-folds, and we scan the hyperparameter space for 25 iterations, choosing the model with the best accuracy. The hyperparameters we experimented with (and their intervals) are: \emph{number of trees} (5--200), \emph{learning rate} (0.0--0.3), \emph{maximum depth of a tree} (6--12), \emph{subsample ratio of the training instances} (0.8--1.0), and \emph{fraction of columns to be subsampled} (0.8--1.0). All those values can be easily adjusted within the code. 

In the following subsections, we explain the system by using a running example with the \emph{red wine quality} physicochemical data set~\cite{Cortez2009Modeling} obtained from the UCI ML repository~\cite{Dua2017}. 
The data set represents a very imbalanced multi-class classification problem and consists of 11 numerical features and 1,599 instances. Consequently, we mapped the six quality categories of wine for the dependent variable to three new classes (similar to Laughter and Omari~\cite{Laughter2020A}) to alleviate this problem. The \emph{fine} class contains 1,319 instances and comprises qualities of 5 and 6. The \emph{superior} quality consists of 217 instances and categories 7 and 8. Finally, the \emph{inferior} class has 63 instances, and it was created by merging qualities 3 and 4.

\subsection{Data Space} \label{sec:data}

In FeatureEnVi, data instances are sorted according to the predicted probability of belonging to the ground truth class, as shown in~\autoref{fig:teaser}(a). The initial step before the exploration of features is to pre-train the XGBoost~\cite{Chen2016XGBoost} on the original pool of features, and then divide the data space into four groups automatically (i.e., \emph{Worst, Bad, Good, and Best}). The vertical black line is the stable threshold anchored precisely at 50\% predictive probability, which separates the correctly from the wrongly classified instances. The other two thresholds partition the prior subspace into their half areas for the default option. However, the user can alter the vertical gray lines as indicated in \autoref{fig:use_case2_sel}(a.1--a.4), with a degree of freedom set to $-$$/$$+$20\% from the defaults. The vertical positioning of the instances is purely used to avoid---as much as possible---overlapping/cluttering issues via jittering. The data space will always be divided into four parts conveying extra information to the user. If no instances belong to a slice of the data space, the system works normally, but there will be no values for the statistical measures (see \autoref{sec:featover}). Overall, the user's goal is to move as many instances as possible from the left side (Worst and Bad subspaces) to the right side (Good and Best subspaces) while avoiding the opposite. Nevertheless, the primary purpose of this view is to provide better local and global explainabilities of the impact of features according to the user-defined slices. In the future, we plan to enable users to determine the number of slices customly (cf.~\autoref{sec:design}).

In our running example, recognizing that the instances are uniformly distributed from \autoref{fig:use_case1_sel}(a.1--a.3), we use the default splitting with 25\% predictive probability intervals. It is essential to focus on the aforementioned slices (especially (a.1) and (a.2)) without ruining the prediction for the correctly classified instances. Once the thresholds are set, we click the \emph{Feature Exploration} button to start our search for features.

\subsection{Feature Selection Techniques} \label{sec:featsel}

\autoref{fig:use_case1_sel}(b) is a table heatmap view with five automatic feature selection techniques, their \emph{Average} contribution, and an \emph{\# Action \#} button to exclude any number of features. As we originally train our ML algorithm with all features, the yellow color (one of the standard colors used for highlighting~\cite{Silva2011Using}) in the last column symbolizes that all features are included in the current phase (if excluded, then B/W stripe patterns appear). The first technique is \emph{Univariate FS} (Feature Selection)~\cite{Jain2021Rank} which uses the ANOVA $F$-$value$ test for selecting the $k$ best features. The value $k$ is always set to the maximum in order to retrieve scores for all features, since we want to avoid removing features automatically (instead, we visualize the scores for the user to decide). The \emph{Impurity-based FI} (Feature Importance) method~\cite{Louppe2013Understanding} is connected to the intrinsic nature of ensemble algorithms to export feature importance scores after their training. Hence, we extract the feature importance scores from the best model we found. \emph{Permutation FI}~\cite{Radivojac2004Feature} is another technique in which the decrease in a model's score is monitored while a single feature value is randomly shuffled~\cite{Breiman2001Random}. The last two techniques are rather similar with one difference: the former method can be biased toward high cardinality features (many unique values) over low cardinality features such as binary features. 
In \emph{Accuracy-based FI}~\cite{Janecek2008On}, we also fit the ML model using one feature at a time and compute the accuracy to evaluate every feature's performance.
Next, as XGBoost~\cite{Chen2016XGBoost} is a nonlinear ML algorithm, we also train a linear classifier (a logistic regression~\cite{Cramer2002Origins} model with the default Scikit-learn's hyperparameters~\cite{Pedregosa2011Scikit}) to compute the coefficients matrix and then use Recursive Feature Elimination (RFE)~\cite{Guyon2002Gene} to rank the features from the best to the worst in terms of contribution.
This technique is referred to as \emph{Ranking-based FS}~\cite{Jeon2020Hybrid} in our VA system. We would like to include further techniques in the future, however, the current selection is specifically assembled to contain one candidate for each of the high-level categories of feature selection methods introduced in~\autoref{sec:intro}. For every method, we normalize the output from 0 to 1 to set a common ground for the user to compare them, as indicated in the legend of \autoref{fig:teaser}(b). Hence, their average is calculated and displayed in the penultimate column. Following the design guidelines from the conventions introduced by prior works~\cite{Lex2010Caleydo,Eisen1998Cluster}, we choose red and green colors for the table heatmap. This view also automatically extends for the newly-generated features from combinations of already existing features (cf. \autoref{fig:teaser}(b)). The original features used for the creation of new features are depicted in dark gray in the last column of the table heatmap view. The table is automatically sorted based on the average; however, Impurity-based FI is selected by the user for the \autoref{fig:teaser}(b)) scenario. Due to this selection, the table heatmap resorts the features from the highest to the lowest importance only according to the XGBoost model's inherent feature importance. More details can be found in~\autoref{sec:featdet}.

At this phase, we want to identify any number of features that can be excluded from the analysis because they contribute only slightly to the final outcome. This will help the system to reduce the computational time needed to train the model—something significant in a real-world scenario. Indeed, if we take a closer look, the last five features underperform and are having a shallow impact on the final result (see \autoref{fig:use_case1_sel}(b)). Thus, we exclude those features one by one with the interactive cells from the \emph{\# Action \#} column. In \autoref{fig:use_case1_sel}(c), we can notice the black-and-white stripe pattern indicating that a feature is excluded. The remaining open question that we will examine in \autoref{sec:featover} and \autoref{sec:featdet} is: shall we continue with the removal of features (e.g., F4 marked with the red ellipsoid shape) or stop at this point?

\subsection{Feature Space Overview} \label{sec:featover}

After the initial removal of features, as described in~\autoref{sec:featsel}, we take a look into the radial tree that presents statistical information about the impact of the currently included features. 
\hl{The core aim of this view is to examine the impact on various subspaces, since removing a feature might appear the right choice globally, but can cause issues for instances in a particular subspace.} 
This hierarchical visualization exploits the connections of these features (see~\autoref{fig:use_case1_sel}(d.1--d.4)) with the four subspaces we defined in~\autoref{sec:data}, which is the \emph{inner layer}. The top part highlighted in a rectangular red box is the whole data space with all the slices (text in bold), and it is currently active (black stroke instead of gray) for the Feature Space Detail panel we will explore in~\autoref{sec:featdet}. One of the purposes of this view is to serve as an \emph{overview} for the graph visualization described in the next subsection (i.e., the more detailed \emph{context}).
The radial tree's gray nodes are the six remaining features that constitute the \emph{middle layer}. Finally, the \emph{outer layer} is used for the feature transformations, as depicted in~\autoref{fig:use_case1_tra}(a). The user can click on any gray circle visible in~\autoref{fig:use_case1_sel}(d.1--d.4) to activate another slice, thus updating the graph visualization (discussed in the next subsection).  

The impact of the features is visually represented here in different ways. Each node is enclosed by a circular bar indicating the absolute value of \emph{Pearson's correlation coefficient} \cite{Pearson1895Correlation} with the dependent variable, from 0\% to $|$100\%$|$. 
The light blue to dark blue color is the \emph{MI} computed between each feature and the target.
The colors of the edges represent the aggregated difference of a feature, if it was transformed at all. 
Green means that the correlation between features decreases with most transformations, while red depicts the increase, respectively. 
With those coherent design choices, the consistency between the radial tree and the graph visualization described in~\autoref{sec:featdet} is preserved, and the users' cognitive load is reduced (see~\autoref{sec:design} for further discussion of the core design choices).

The radial tree representation and the graph visualization (see below) use a layout panel visible at the bottom of~\autoref{fig:teaser}(c). It supports several interactions such as \emph{zooming}, \emph{rotating}, \emph{returning to the initial view}, and expanding/retracting particular slices with the \emph{Toggle} options. Zooming enables users to compare effortlessly the statistical measures computed for the different subspaces depicted in the radial tree representation. Rotating allows users to read the text labels easily. 
Returning to the initial view can also be a helpful shortcut during the interactive exploration. 
The toggle options assist users in collapsing unimportant data subspaces in particular cases or vice versa. Furthermore, this functionality facilitates FeatureEnVi to scale for data sets with many more features (see~\autoref{sec:limit} for the respective discussion). 


Next, we focus on the overall inspection of features for all instances (see \autoref{fig:use_case1_sel}(d.1--d.4)). 
F4 (the ellipsoid shape) appears the worst in terms of target correlation (the small circular bar), and it has one of the lowest MI values (light blue color). 
The impact is also the weakest for the Worst and the Best subspaces compared to the other features. 
However, for the remaining slices, it has a higher impact than other features, as can be seen in \autoref{fig:use_case1_sel}(d.2 and d.3).
The detailed data space view should thus be examined next. 

\subsection{Feature Space Detail} \label{sec:featdet}

The graph visualization in~\autoref{fig:use_case1_sel}(e) uses glyphs to encode further details for every feature, i.e., \emph{node}.
The per target correlation and MI are the same as in the radial tree, but an additional horizontal bar chart encodes the \emph{per class correlation} of each feature with three colors (olive for fine class, purple for superior class, and turquoise for inferior class). Moreover, there are four states that indicate if the \emph{VIF} per feature was greater than 10, between 10 and 5, between 5 and 2.5, and finally, less than 2.5.
They are represented by up to four circular segments in gray (see \autoref{fig:use_case1_sel}(e)) laid out clockwise within the node. 
We decided for these states as the prior research suggests they reflect concerns or problematic cases of colinearity~\cite{Vittinghoff2011Regression,Menard2002Applied,Johnston2018Confounding}.
\emph{Edges} are displayed between feature nodes with the correlation above the current threshold (0.6 in our example), with edge widths encoding correlation values.

\begin{table}[tb]
  \captionsetup{justification=centering, labelsep=newline}
  \centering
  \caption{Summary of the Currently Supported Feature Transformations}
  \label{tab:trans}
  \includegraphics[width=\columnwidth]{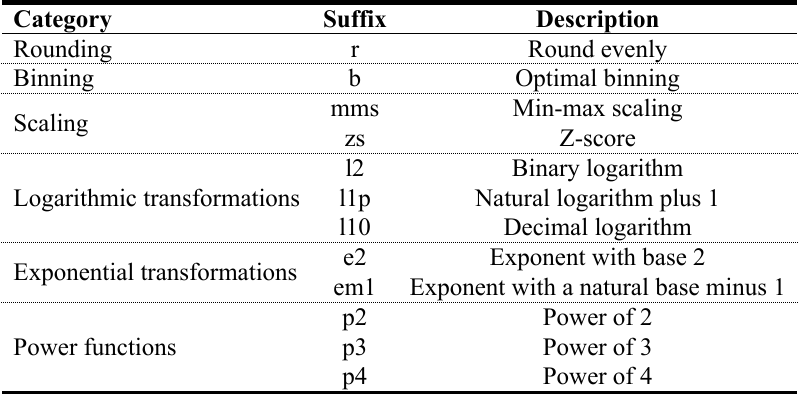}
\end{table}

 \begin{figure*}[tb]
  \centering
  \includegraphics[width=\linewidth]{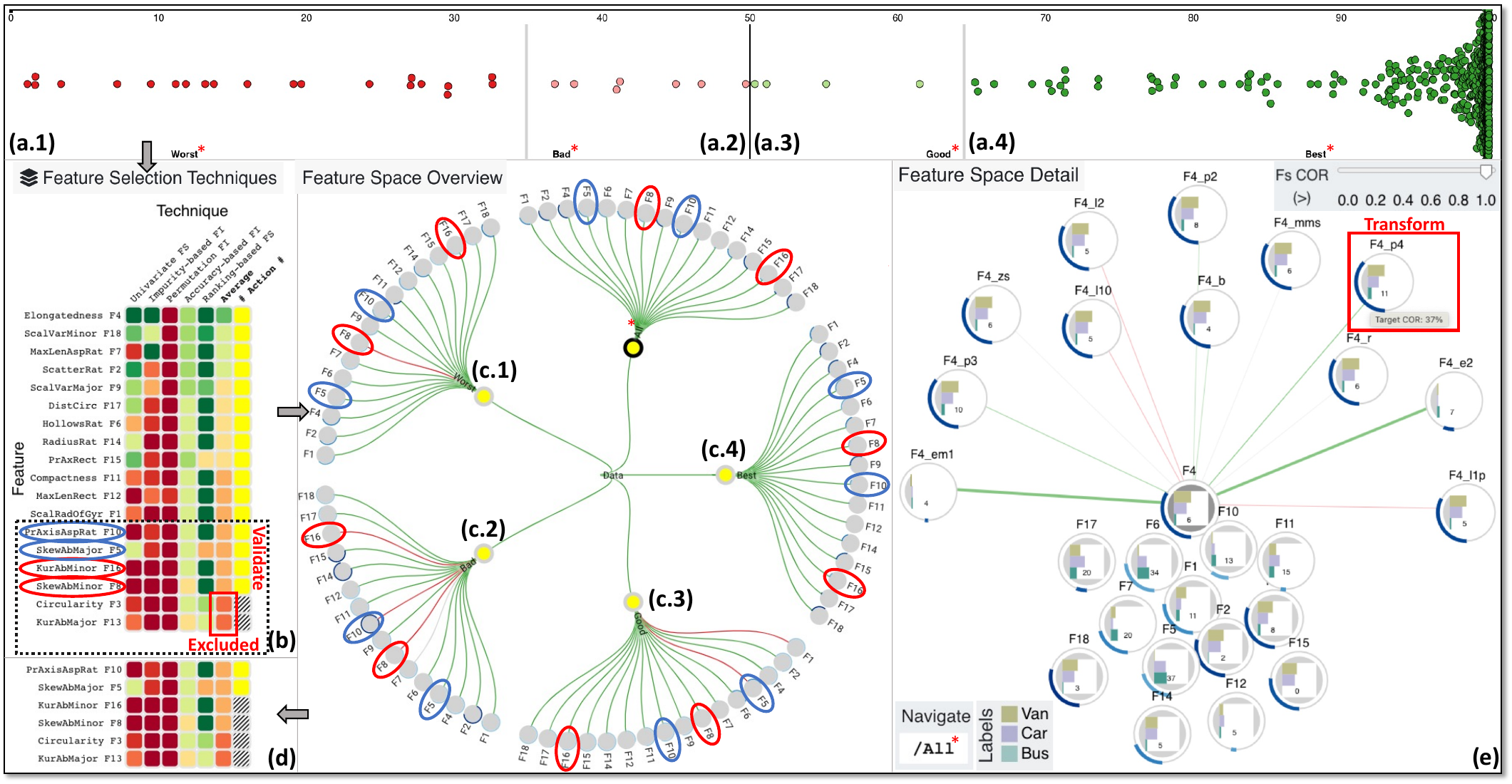} \vspace{-5mm}
  \caption{The process of features' exploration in a vehicle recognition scenario. (a.1) to (a.4) depict the change of the thresholds for the different data slices to intensify the responses for borderline instances. In (b), the user excluded unimportant features and then validates the remaining features through the radial tree visualization. (c.1) to (c.4) contain the main statistical measures (target correlation and mutual information) and allow the user to discover that F8 and F16 should be excluded, as shown in (d). Finally, (e) illustrates the transformation alternatives of the most contributing feature (i.e., F4); the user employs a power function since it increases the correlation with the target and reduces slightly the correlation between features.}
  \label{fig:use_case2_sel}
\end{figure*}

The graph visualization uses a force-based layout algorithm, and we allow the user to select different timestamps of the graph generated by a dynamic optimization approach~\cite{Gove2018It}. Whenever a node in focus is not clearly visible, the user can choose another pre-stored layout to reduce clutter. We tweaked the parameters to put more weight to the avoidance of edge crossings ($\times$1.00), then the deviation of angular resolution ($\times$0.75), next the minimum angular resolution ($\times$0.40), and finally the edge crossing angle ($\times$0.10) in the graph visualization. We found empirically that these values produce good graph layouts that match the goal of our tool.
The \emph{weighted graph readability widget} is located at the bottom of the panel in~\autoref{fig:teaser}(c), which also includes an edge filtering slider based on minimum \emph{features' correlation} (Fs COR) value. 

In the panel below the graph view (see \autoref{fig:teaser}(d)), the user is able to select between two modes: (1) \emph{Feature Transformation} and (2) \emph{Feature Generation}. 
The user is also informed about the currently navigated data space and provided with static legends describing this view.
When the \emph{Feature Transformation} mode is active, clicking on a node will expand it, exposing all supported transformations which are based on mathematical functions (\autoref{fig:use_case1_tra}(a)).
The available transformations can be divided into six categories listed in~\autoref{tab:trans}. 
For \autoref{fig:use_case1_tra}(a and b), the red and green colors have the same meaning as explained in \autoref{sec:featover}. 
The magnitude of increase/decrease is mapped to the line width in \autoref{fig:use_case1_tra}(b). 
For \emph{Feature Generation}, FeatureEnVi supports four basic arithmetical operations ($+, -, \times, /$). 
The selected features are highlighted in the dark gray color (because it matches the default color, which is gray) of the VIF metric's region, as demonstrated in~\autoref{fig:teaser}(d), and the combinations are generated for the two or three selected features automatically, as can be seen in~\autoref{fig:teaser}(b). It is up to the user to select the best generation; however, he/she is being assisted by the automated selection techniques, as described in \autoref{sec:featsel}.
In the future, we plan to support custom transformation and generation of features (see~\autoref{sec:eval}).

\hl{By comparing the lengths of the circular bars in \autoref{fig:use_case1_sel}(e), we see that the lowest overall target correlation is reported for F4 (on hover shown as 10\%).}
Also, the MI exhibits low values for both F3 and F4. As we proceed, we observe that F3, F4, and F6 may cause problems regarding collinearity based on the VIF heuristic (2 out of 4 pieces of the symbol). Additionally, the per-class correlation is 12\% for the fine label, and weak for the other classes. Finally, the correlation between F4 and F6 (or F4 and F3) is above 0.6; thus, it can be considered moderate to strong. 
This leads us to excluding F4, with the results in \autoref{fig:use_case1_tra}(b) being promising since the remaining features (in the red box) show no signs of strong correlation with each other or any collinearity if we consider VIF. 

From~\autoref{fig:teaser}(d), we can already see the transformations we applied in the different features with the suffixes attached. At this stage, we choose the feature generation mode and pick three features (F1\_l2, F6\_b, and F9\_l10) that correlate in a similar way with the individual classes. Moreover, F1\_l2 is moderately correlated with F6\_b. The selected nodes are highlighted with dark gray color, and the combinations of the different mathematical operations are prepared dynamically (see \autoref{fig:teaser}(b)). From this subfigure, we spot that F1\_l2$\times$F6\_b contributes much more than the source features if we sort the table according to the Impurity-based FI. Therefore, we add this new feature.

\subsection{Process Tracker and Predictive Results} \label{sec:results}

Each action of the user is registered in the punchcard visualization (see \autoref{fig:teaser}(e)). 
The basic recorded steps are: (1) \emph{Include}, (2) \emph{Exclude}, (3) \emph{Transform}, and (4) \emph{Generate} for each feature. 
The size of the circle encodes the order of the main actions, with larger radii for recent steps. 
The brown color is used only if the overall performance increases. 
The calculation is according to three validation metrics after we subtract their standard deviations.
The grouped bar chart presents the performance based on accuracy, weighted precision, and weighted recall and their standard deviations due to cross-validation (error margins in black). 
Teal color encodes the current action's score, and brown the best result reached so far. The choice of colors was made deliberately because they complement each other, and the former denotes the current action since it is brighter than the latter. 
If the list of features is long, the user can scroll this view. 
Finally, there is a button to \emph{extract} the best combination of features (i.e., the ``new'' data set).

\hl{The brown circles in the punchcard in~\autoref{fig:teaser}(e) enable us to acknowledge that the feature generation boosted the overall performance of the classifier.}
High scores were reached in terms of accuracy, precision, and recall. \hl{\textbf{All in all}} with FeatureEnVi, we improve the total combined score by using 6 well-engineered features instead of the original 11. 
On the contrary, Rojo et al.~\cite{Rojo2020GaCoVi} reported a slight decrease in performance when selecting 6 features for this task as a regression problem.

\begin{figure*}[tb]
  \centering
  \includegraphics[width=\linewidth]{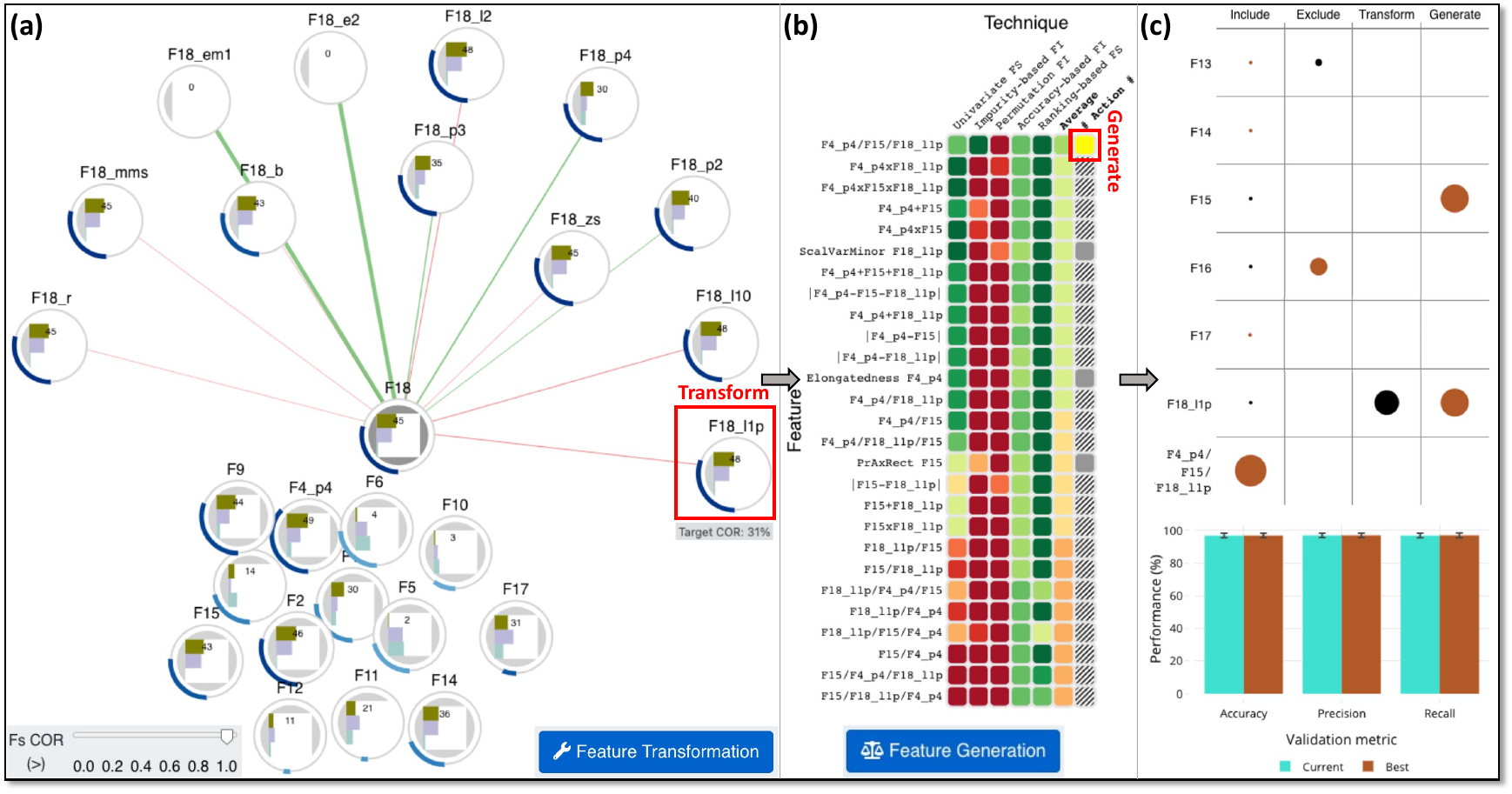} \vspace{-5mm}
  \caption{
  Further transformations and generation of new features for the vehicle data. 
  (a) presents another transformation of the second most impactful feature (according to \autoref{fig:use_case2_sel}(b)). F4\_p4$/$F15$/$F18\_l1p is the most important combination (see the darker green color in (b)). The punchcard visualization in (c) indicates that when we removed F16, the performance increased and that the new feature boosted, even more, the predictive results. For all metrics in the grouped bar chart, the best values are equal to the current results.}
  \label{fig:use_case2_tra}
\end{figure*}

\subsection{Use Case: Understanding Feature Engineering} \label{sec:useCase}

%
We now describe a different case of 
how FeatureEnVi could be used to explore the feature space and engineer 18 features for the multi-class \emph{vehicle silhouettes} data set~\cite{Siebert1987Vehicle}, comprising vehicle data assigned to one of three classes: \emph{van}, \emph{car} or \emph{bus}.
The class distribution is rather imbalanced, with 199 vans, 429 cars, and 218 buses.
For this use case, we use the same ML algorithm, hyperparameter optimization method, and cross-validation strategy as in the previous application. 
This use case was performed by us, and it was the first time we explored this particular data set.

\textbf{Slicing the data space and selecting subsets of features.} 
Similar to the workflow described above, 
we start by choosing the appropriate thresholds for slicing the data space. As we want to concentrate more on the instances that are close to being predicted correctly, we move the left gray line from 25\% to 35\% (see \autoref{fig:use_case2_sel}(a.1 and a.2)). This makes the Bad slice much shorter. Similarly, we focus more on the correctly-classified instances by changing the position of the right gray line from 75\% to 65\% (cf. \autoref{fig:use_case2_sel}(a.3 and a.4)). From the table heatmap view in \autoref{fig:use_case2_sel}(b), we realize that F13 and then F3 can be excluded from the features list. For the remaining features, we have to validate our hypotheses through the statistical measures of the radial tree visualization. We check how F8, F16, F5, and F10 perform in various data subspaces, as shown in \autoref{fig:use_case2_sel}(c.1--c.4). All these features have rather low MI in the \emph{All} space due to light blue color. Hence, the difference is mainly in the linear correlation of those features with the dependent variable. F8 appears the least correlated with the target variable (small circular bar). F16 is similar to F8 regarding correlation, except for the Good subspace in~\autoref{fig:use_case2_sel}(c.3). 
Thus, these two features should be removed. 
However, F5 and F10 seem more important for all instances, e.g., they both have a strong correlation with the class label in the Bad subspace in~\autoref{fig:use_case2_sel}(c.2). 
The result of this investigation can be seen in~\autoref{fig:use_case2_sel}(d).

\textbf{Transforming the most impactful features.} 
After the feature selection phase, we use the graph view to transform the most contributing features (F4 in~\autoref{fig:use_case2_sel}(e) and F18 in~\autoref{fig:use_case2_tra}(a)).
\hl{The car class could be the easiest to classify as it contains the most instances, thus we try improving correlation with other classes.}
For F4, the power function with exponent 4 almost doubles (from 6 to 11) the correlation for the bus class.
The logarithmic transformation for F18\_l1p in~\autoref{fig:use_case2_tra}(a) boosts the correlation with the van class from 45\% to 48\%.
For both features, the overall correlation with the target class increases with these transformations, while the others produce worse results, so we avoid them.

\textbf{Generating features from similarly behaving features.} 
The newly-transformed F4\_p4 and F18\_l1p are both powerful for the first two classes. 
For the same reason, F2, F9, or F15 could be good candidates for blending with the above features (see \autoref{fig:use_case2_tra}(a)).
We pick one (in this case, F15) and generate a new feature by automatically mixing these features with the operations described in~\autoref{sec:featdet}. 
The best sequence that was identified is F4\_p4$/$F15$/$F18\_l1p. Based on the feature selection techniques, five combinations are more impactful than the three source features. At this point, we activate the new feature being at the very top of the table heatmap. Thus, this feature is being added to the original list of features dynamically (cf. \autoref{fig:use_case2_tra}(b)). 

\textbf{Tracking the process and evaluating the results.} 
To verify each of our interactions, we continuously monitor the process through the punchcard, as shown in \autoref{fig:use_case2_tra}(c). From this visualization, we acknowledge that when F16 was excluded, we reached a better result. The feature generation process (described previously) led to the best predictive result we managed to accomplish. The new feature is appended at the end of the list in the punchcard. From the grouped bar chart in \autoref{fig:use_case2_tra}(c), the improvement is prominent for all validation metrics because the brown-colored bars are at the same level as the teal bars. \hl{\textbf{To summarize,}} FeatureEnVi supported the exploration of valuable features and offered transparency to the process of feature engineering. The following case study is another proof of this concept.

\section{Case Study} \label{sec:case}%
  \begin{figure*}[tb]
  \centering
  \includegraphics[width=\linewidth]{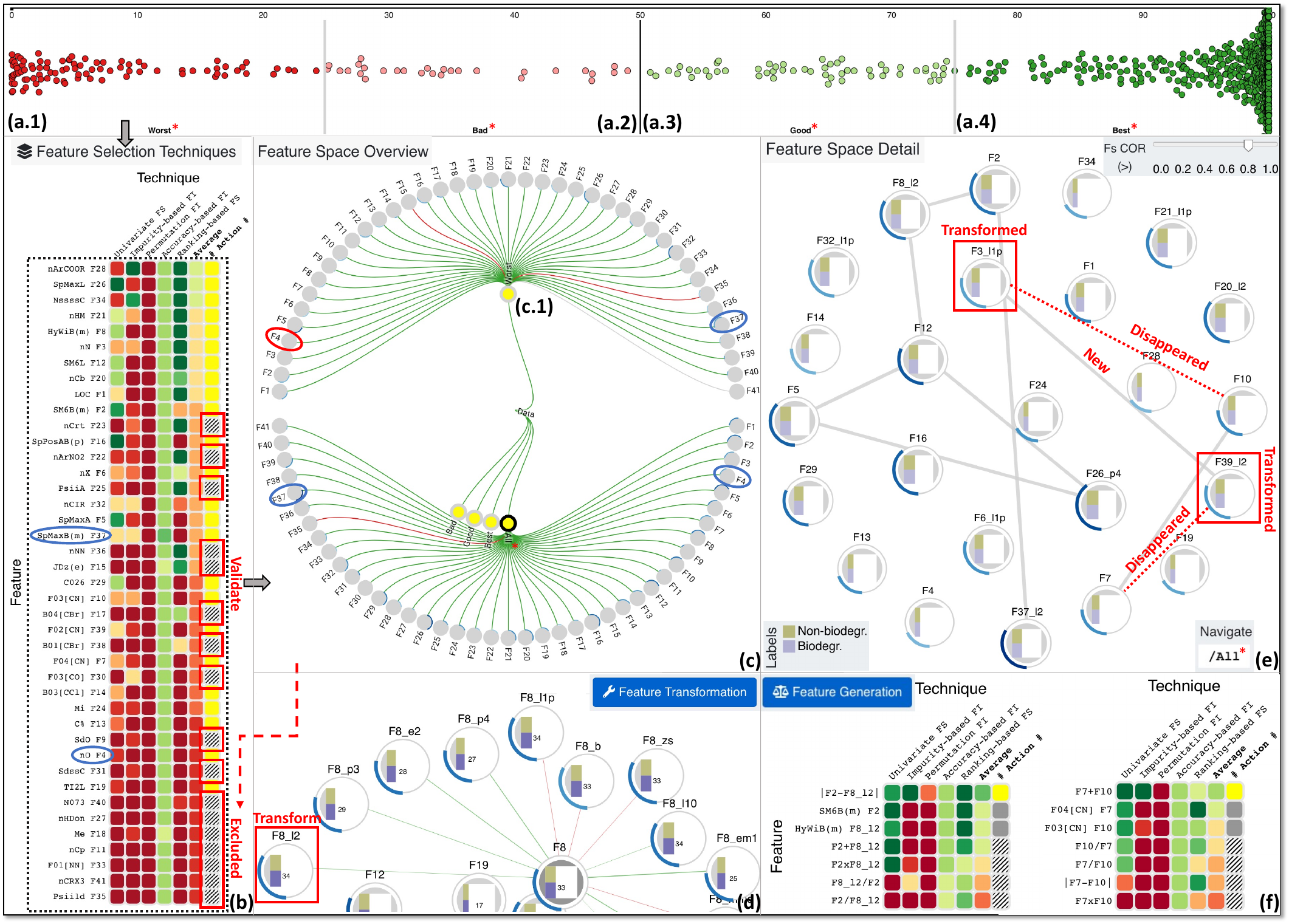} \vspace{-5mm}
  \caption{Engineering features for improved predictive performance. From the pre-training phase, we detect that most of the instances belong to the Best slice (a.4), then the Worst slice (a.1), followed by the remaining slices (a.3 and a.2). In view (b), we validate every feature by working in synergy with the table heatmap and the radial tree shown in (c). We pick the whole data space to be juxtaposed against the Worst slice since it contains many wrongly classified instances (see c.1). Interestingly, F4 appears unimportant for the Worst slice, but it is somewhat correlated with the target variable in the All space. Later, we explored this particular feature in the Bad slice, and we confirmed that it is valuable there. Consequently, we decided that it should be retained. F37 is an important feature for both spaces under investigation, hence, it is also kept in the pool of features. \hl{In (d), we observed that the logarithmic transformations are the most impactful in terms of increase in target correlation, but at the same time, F8\_l2 and F8\_l10 are the only features that reduce the correlation between features. Thus, we choose one of the best candidates.} After the transformations are applied in different features, F3\_l1p and F39\_l2 are connected due to the new logarithmic transformations, as seen in (e). The between these features correlation has been increased compared to before. However, the links between F3\_l1p with F10 and F39\_l2 with F7 have disappeared because the transformations altered the instances' values. Finally, we experimented with feature generation by selecting strongly correlated features, and we discovered that $|F2-F8\_l2|$ and $F7+F10$ combinations are better than their stock versions, as seen in (f).}
  \label{fig:use_case3}
\end{figure*}

In addition to the use cases discussed above, we present a case study with a focus on helping users to understand the feature engineering process and improve the classification results. The case study also provides useful insights that can be used to improve our visualization approach in the future.


This case study was carried out together with
an invited ML expert (PhD in \emph{computer science}, currently \emph{associate professor} in \emph{artificial intelligence}, with \emph{20} years of experience in ML, but no prior experience in VA). 
The expert collaborated with us using FeatureEnVi to improve the results of a prior study about the relationships between \emph{chemical structure and biodegradation of molecules}, described in the work by Mansouri et al.~\cite{Mansouri2013Quantitative}.
The \emph{QSAR (Quantitative Structure Activity Relationships) biodegradation} data set represents a binary classification problem where molecules are assigned to either the \emph{biodegradable} or \emph{non-biodegradable} classes. The class distribution is relatively imbalanced, with 284 degradable and 553 non-degradable molecules for the training set containing 41 different features. The test set contains 72 and 146 instances in each class. At the same time, the external validation set consists of 191 and 479 instances, respectively.
For their solution, the authors trained three ML models (KNN~\cite{Fix1989Discriminatory}, PLSDA~\cite{Brereton2014Partial}, and SVM~\cite{Hearst1998Support}) and then combined their results using two consensuses. We worked hand-in-hand with FeatureEnVi to surpass their predictive results by solely manipulating the features of this data set. Only for this case study, the cross-validation strategy was adapted to 5-folds instead of 8-folds for comparison reasons with Mansouri et al.~\cite{Mansouri2013Quantitative}.

\textbf{Selection of impactful features for the Worst subspace.} 
We began our investigation by examining the distribution of instances in the explorable subspaces. We noticed that most instances are correctly classified with more than 75\% predicted probability (i.e., high confidence), as shown in~\autoref{fig:use_case3}(a.4). The invited ML expert found the 25\% predicted probability intervals a considerably rational choice. Thus, it was adopted as the default threshold strategy for this analysis. The focus was on the Worst slice (cf.~\autoref{fig:use_case3}(a.1)) because it contained many instances that could be hard to classify. These instances could lead to low generalizability and hurt the overall predictions since such rare cases can occur in the test and external validation sets, and it is usually not trivial to classify them correctly.
At this point, we worked with the table heatmap view (\autoref{fig:use_case3}(b)) to validate all features and exclude a subset of features. The automatic feature selection techniques cannot be trusted when left alone due to the contradictions arising in various situations. Therefore, we utilized the statistical measures visible in the radial tree (see~\autoref{fig:use_case3}(c)).
The radial tree had three collapsed data subspaces (a.2--a.4) except for All and Worst subspaces. We performed this action because there are too many features to be explored at once, and FeatureEnVi provides this capability to alter the layouts in order to scale for high-dimensional data sets. Basically, the core statistical measure to examine in the radial tree is the target correlation depicted as a circular bar of different sizes for each feature in every data subspace. Moreover, a supportive measurement is the MI per feature that should have a light blue color in the cases of a potentially removable feature. 
Going from the bottom to the top of the feature importance list sorted by the Average importance, we excluded all features with a target correlation value greater than 15\% according to the All space. Two interesting examples of features are F4 and F37. The former was powerful only for the Accuracy-based FI and slightly better for the Univariate FS than the remaining feature selection techniques. The latter was moderately good for all techniques except for Permutation FI and Ranking-based FS. 
We concentrated on the conjunction of those automatic approaches with the statistical measures offered by FeatureEnVi. Specifically, F4 resembled an unimportant feature for the Worst subspace, as shown in~\autoref{fig:use_case3}(c.1). Albeit that, when closely explored in the whole data space, it was more impactful than other features (with a target correlation value of more than 15\%). Afterwards, we collapsed the Worst slice and expanded the Bad slice to explore the impact of the feature. Indeed in that particular slice, this feature appeared to be rather important (not shown due to space limits). These recursive actions reassured us that F4 should be kept in the pool of features. On the other hand, F37 was very impactful in both All space and Worst subspace (cf.~\autoref{fig:use_case3}(c and c.1)), hence, it remained in the list of active features. 
Without the support of additional statistical measurements such as target correlation, we could have made possible mistakes because automatic feature selection techniques sometimes contradict each other, and they are not fully trustworthy in certain situations. After this phase, 24 out of the 41 original features remained in use.

\textbf{Transformation of features with guided decisions.}
For the feature transformation phase, the first step was to analyze every feature with a top-down approach according to the sorting implied in the table heatmap (\autoref{fig:use_case3}(b)). FeatureEnVi facilitates the exploration of multiple contradictory criteria again, even while choosing the individual transformations. In~\autoref{fig:use_case3}(d), we acknowledged a case where, according to the target correlation measure, all logarithmic transformations were sufficient. \hl{However, F8\_l2 and F8\_l10 were better candidates for transformation since the correlation between features decreased (green lines instead of red-colored) when choosing these options against the F8\_l1p transformation.} All performed transformations are displayed in~\autoref{fig:use_case3}(e). 
F26 was the only feature transformed using a power function, while F21, F8, F3, F20, F6, F32, F37, and F39 were transformed with logarithmic transformations. The rest of the features remained stable. 
During this exploration process, we applied the following rule: 
transformation of not so impactful features with extremely low target correlation value (less or equal to 20\%) was skipped. 
As mentioned earlier in~\autoref{sec:system}, most useful transformations appear in impactful features. We noticed that when one feature of a pair of strongly correlated features gets transformed, then there is a possibility that their correlation will decrease. The opposite effect is also plausible. Consequently, not all features that are strongly correlated with each other should be transformed, which is another gained insight. 
In~\autoref{fig:use_case3}(e), we transformed F3\_l1p and chose to keep the same F10 to remove the initial link between these two features. The same approach was followed for features F39\_l2 and F7, which led to an identical outcome. These transformations, though, created a new link of strongly correlated features with F3\_l1p being connected to F39\_l2 (see \autoref{fig:use_case3}(e)). With the aforementioned rule and insight in mind, we managed to decrease the number of correlated features with more than 0.8 Fs COR to 8 connections (shown in gray lines) instead of the initial 10 links. Also, the VIF was reduced from greater than 5 to greater than 2.5 (i.e., from 3 to 2 symbols active) for F10 compared to before the transformation phase. Similarly, F26, F21, F20, F16, and F13 went from greater than 10 to greater than 5 in VIF (from 4 to 3 symbols active), while F4's VIF had a bigger difference gap from greater than 10 to just greater than 2.5 (i.e., from 4 to 2 symbols active).
The increase in target correlations along with decreases in-between feature correlations and VIF were the main steps to reach a higher performance alongside the steps discussed next. 
At this point, 9 out of the 24 features were transformed.

\textbf{Generation of features from correlated pairs of features.}
At this stage, we concentrated on generating new and more effective features from the existing ones. FeatureEnVi provides assistance in this procedure by highlighting only the strong correlation of features (with the current value for the Fs COR) in gray lines (cf. \autoref{fig:use_case3}(e)). We tested all old links between pairs of features that are visible in (e) because they are computed quicker than the examination of three features concurrently. The new connection was omitted because it was generated artificially from our interactions. In detail, F37\_l2 with F2, F8\_l2 with F12, F8\_l2 with F2, and finally, F7 with F10 were explored a pair at a time through the updatable table heatmap view. The first two pairs of features, when combined, gave only less important features compared to the original features used to create them. Thus, we did not include any new features from these pairs. On the other hand, F8\_l2 and F2 when combined with the $|F2-F8\_l2|$ mathematical operation, generated a more impactful feature (see \autoref{fig:use_case3}(f)). The same is true for the F7 and F10 pair of features when using the $F7+F10$ operation. To sum up, two new features were generated, resulting in 26 well-engineered features instead of the 41 original features.

\textbf{Evaluation with the test and external validation sets.} 
Throughout the aforementioned phases, we utilized feature engineering to improve the most powerful XGBoost model found through hyperparameter tuning. To verify whether our findings were reliable, we applied the resulting ML model to the same test and external validation sets as Mansouri et al.~\cite{Mansouri2013Quantitative}, see \autoref{results}. For the test data set, the reported accuracy was approximately 87\%.
In our case, we reached 89\% for accuracy after the feature engineering process was completed.
Using our approach, we managed to achieve the same accuracy as before, 89\%, compared to 83\% reported by Mansouri et al.~\cite{Mansouri2013Quantitative} for the additional external data set. For precision and recall, we always use \emph{macro-average}, which is identical to Mansouri et al.~\cite{Mansouri2013Quantitative}. On the one hand, the precision was 4\% lower in both test and external validation sets for our analysis. On the other hand, the recall was 5\% higher for the test set and 9\% higher for the external validation data set, which improves the generalizability of the ML model for similar data. 
\hl{\textbf{In summary,}} our system not only supports the exploration of many different perspectives of the complex relationships between features, but also is capable of optimizing features (and consequently the underlying ML model) \hl{to efficiently use fewer well-engineered features} and boost the predictive performance.

\begin{table}[tb]\centering
\captionsetup{justification=centering, labelsep=newline}
\begin{threeparttable}
\caption{Results for the Test and External Validation Sets\\for the Case Study}
\label{results}
\setlength\tabcolsep{0pt} 
\begin{tabular*}{\columnwidth}{@{\extracolsep{\fill}} lcccc}
\toprule
\multirow{2}[3]{*}{\shortstack{Validation \\ Metric}} & \multicolumn{2}{c}{FeatureEnVi} & \multicolumn{2}{c}{Mansouri et al.~\cite{Mansouri2013Quantitative}} \\
\cmidrule(lr){2-3} \cmidrule(lr){4-5}
 & Test & External Validation & Test & External Validation \\
\midrule
     Accuracy & 89\% (+2\%) & 89\% & 87\% & 83\% \\
     Precision & 88\% (+2\%) & 87\% & 92\% & 91\% \\
     Recall & 87\% (+3\%) & 85\% & 82\% & 76\% \\
\bottomrule
\end{tabular*}
\smallskip
\scriptsize
\footnotesize{\hl{The value in parentheses (+X\%, where X is a number) highlights the increase in predictive performance due to the use of FeatureEnVi.}}\\
\end{threeparttable}
\end{table} 

\section{Evaluation} \label{sec:eval}%
Following the guidelines from prior works~\cite{Ma2020Explaining,Ming2020ProtoSteer,Xu2019EnsembleLens,Chatzimparmpas2021StackGenVis}, we conducted online semi-structured interviews with three experts to collect qualitative feedback about our system's effectiveness.
The first ML expert (\textbf{E1}) is a senior lecturer in mathematics with a PhD in this field. 
He has approximately 3.5 years of experience with ML, and he currently works with reinforcement learning. 
The second ML expert (\textbf{E2}) is a senior researcher in software engineering and applied ML working in a governmental research institute as an adjunct professor. He has worked with ML for the past 8 years. The third expert (\textbf{E3}) is an associate professor in computer science and a visualization researcher, with 15 years of experience in ML and VA. 
The latter two experts have PhDs in computer science; no expert reported any colorblindness issues. 
Each interview lasted about 1 hour, and the interview procedure was: 
(1) presentation of the key goals of FeatureEnVi, (2) demonstration of the functionality of each view and experts' interaction with the system using the \emph{iris flower} data set~\cite{Dua2017}, and (3) explanation of the process of reaching the results for the \emph{red wine quality} use case in~\autoref{sec:system}.
The first part serves as an introduction to the participants to understand the research problems that FeatureEnVi strives to address. 
The second part focuses on the overall usability as well as the interactions and the connections between views with a simple and well-known data set. 
The third part is about assessing the system's effectiveness and getting familiar with the workflow.
The participants were free to comment on anything, and their main points can be found below. 

\textbf{Workflow.}
\textbf{All experts} commented that the workflow of FeatureEnVi is straightforward, because it is mainly linear despite involving optional iterative steps. \textbf{E2} stated that feature engineering is usually very time consuming, especially without the support of a system like ours. \textbf{E3} also agreed with us that the features have an important influence on the predictive model-based quality and affect the generalization ability of the final ML model. Furthermore, he noticed that it is difficult to judge how each feature should be engineered when there is contradicting statistical evidence. Without FeatureEnVi, it would have been risky to make a deterministic decision. The connection between the features present in the  radial tree visualization and the instances' reallocation in the data space at the top of FeatureEnVi helps to identify impactful features (as highlighted by \textbf{E1}). 
Hence, it is up to users to understand which features matter more for the subspaces (locally) and the entire data space (globally), which provides transparency and enhances the trustworthiness of the feature engineering process, as outlined by \textbf{E1} and \textbf{E2}. \hl{Although FeatureEnVi works better with a limited number of features (e.g., 41 for the case study in~\autoref{sec:case}), \textbf{E2} suggested that a prephase with an AutoML system~\cite{Swearingen2017ATM} or a DR algorithm~\cite{Espadoto2021Toward} could be a solution, if used to set specific boundaries by investigating the relations between features. 
The ultimate goal is to exclude non-contributing features prior to extensive analyses within FeatureEnVi.}
Finally, \textbf{E3} discussed a recent example application in design materials~\cite{Moosavi2020Role} that he worked with, which could benefit from our tool. In particular, he said: ``automatic ML can combine materials (i.e., features) without them having any physical meaning or being possible to exist; all that just for the sake of improving the predictive accuracy''.

\textbf{Visualization and interaction.}
\textbf{E1} and \textbf{E2} were surprised by the promising results we managed to achieve with the assistance of our VA system in the \emph{red wine quality} use case of~\autoref{sec:system}. Initially, \textbf{E1} was slightly overwhelmed by the number of statistical measures mapped in the system's glyphs. However, after the interview session was over, he acknowledged that FeatureEnVi assisted him with hover interactions, and he even requested more metrics. He stated: ``the trick here is to concentrate on specific measurements at a time for multiple features''. \textbf{E2} was particularly enthusiastic about how simple and informative the visualization of the data space is. \hl{\textbf{E1} also added that the radial tree visual metaphor encourages him to engage further with the tool since it follows a structural design found in nature. Formulating and testing hypotheses about features is possible through this view due to the several implemented interaction mechanisms, as noted by~\textbf{E1}. Moreover, \textbf{E1} and \textbf{E3} appreciated the tool's ability to return back to a previous step, if necessary. The former was positively surprised by the ease of extracting the engineered features, and he specifically stated: ``this is awesome''.} The latter suggested an alternative option for the data space visualization which could have been to aggregate the instances belonging to the different subspaces and use a histogram instead of the beeswarm plot. Another choice is a Sankey diagram which will make the view more complex than it currently is, but it can capture all the steps in between. Both visual representations are worthy options; however, this view has a two-fold purpose: we allow users to adjust the thresholds for the slices directly and, at the same time, we enable them to observe the instances' movement between different subspaces (while reducing cluttering with the use of the available vertical space). The same expert provided us with a further extension that can improve the efficacy of the beeswarm plot, which is described in the following paragraph. He also mentioned that a confusion matrix for visualizing the validation results might be a more detailed approach. Despite that, the bar chart fits the available space, and it scales better in multi-class classification problems. We plan to perform a user study to test diverse visualizations.
\hl{We also intend to extend the support for user customization, e.g., to hide some of the statistical measurements if preferred by the user.}

\textbf{Limitations.}
\textbf{E1} and \textbf{E2} were concerned about the \emph{scalability and efficiency} of the system. 
Nevertheless, the use of automatic feature selection techniques as one of the first workflow steps is designed exactly for such purposes.
Another possible improvement is to utilize parallel processing on powerful cloud servers.
Progressive VA and data science workflows~\cite{Stolper2014Progressive,Turkay2018Progressive} could also be effective. \hl{Moreover, \emph{alternative feature selection techniques} for computing feature importance could be incorporated in our tool (e.g., SHAP~\cite{Lundberg2017A}).}
A customized beeswarm plot could facilitate selecting groups of instances and then \emph{explaining why some instances migrated}. 
DR methods could also be helpful here, as noted by \textbf{E3}. Also, he proposed to include \emph{additional filtering options} for all metrics.
The punchcard visualization could allow users to \emph{return at any step and follow another path} (\textbf{E2}). Finally, \textbf{E1} mentioned that it could be useful if our system supported \emph{custom transformation and generation of features} for users to experiment with. We intend to implement the above functionalities.

\textbf{Overall assessment.}
\hl{All experts were pleased with the effectiveness of FeatureEnVi in testing hypotheses about individual features or their combinations. 
Overall, \textbf{E2} was impressed with FeatureEnVi and mentioned: ``I and colleagues would love to try it out with our own data''. 
Also, \textbf{E1} said: ``it is so much fun using your tool''. Afterwards, he continued: ``FeatureEnVi has too much value, even using it in a subset of your data, since you can still generate new knowledge''. \textbf{In conclusion}, the experts deemed FeatureEnVi a useful approach for supporting feature engineering due to the combination of human and computer intelligence.}

\section{Discussion} \label{sec:disc}%
In this section, we discuss several aspects of the design choices as well as limitations of our approach and the current VA system.

\subsection{Design Choices} \label{sec:design}
Here, we extend our discussion on the core design principles of our approach that were introduced in~\autoref{sec:system}.

\textbf{Ground truth versus per class predicted probability.}
In~\autoref{sec:data}, we explain that the data space view presents the predicted probabilities for the ground truth class. A different approach could have been to visualize the predicted probability for every class. Although this idea appears valuable and straightforward for binary classification problems, it will not scale well with multiclass problems addressed by our VA system. It will be tough to present confusion between classes when more than a couple of class labels are available, which is typical in multiclass problems. Moreover, the limited amount of space in that view is another reason we abandoned this idea, despite being a valid alternative. 
While the local explainability offered by this data space view is valuable, the focus of our tool is on the feature engineering process. 

\textbf{Custom subspace slicing.}
Perception and cognition problems could emerge as the number of slices increases. We believe that four slices are already a good start to explore the vast majority of the data space because users will often focus on particular areas of interest.
The two interactive thresholds are a key component here because they allow users to choose the size of each subspace depending on the problem. Additionally, the experts in~\autoref{sec:eval} did not miss the custom slicing functionality. However, FeatureEnVi could be improved further by enabling users to select a custom number of slices, if that is preferred in certain scenarios.

\textbf{Radial tree versus other visual representations.}
The radial tree is an essential supplementary view of the graph visualization because it serves as a contextual anchor. The visual encodings remain the same, so the perception for users does not change overall. This design decision drastically reduces the cognitive load since users do not have to learn  additional encodings from scratch. A potential alternative visual metaphor could have been a tabular form, e.g., an icicle plot. The benefit of this metaphor is the compact space that it captures in comparison to a radial tree. However, the numerous interactions available (e.g., the ability to collapse unimportant slices/central nodes) enable this view to scale well even with more than 40 features. A detailed list of possible layout modifications and interactions can be found in~\autoref{sec:featover}. Despite these functionalities, there is still a bottleneck in the number of features that FeatureEnVi allows users to explore due to visualizations and computational burdens. The first limitation in~\autoref{sec:limit} provides further information on this matter.

\subsection{Additional Limitations and Future Work} \label{sec:limit}
Next we focus on additional limitations identified for our approach, which also hint at possible future improvements. 

\textbf{Scalability for large number of features.}
The limits of FeatureEnVi have already been put to the test with the case study in~\autoref{sec:case}. 
The used data set contains 41 features, which unfortunately are difficult to explore in multiple subspaces simultaneously. Despite that, the invited ML expert was able to monitor the global impact of features along with the local region (Worst subspace) that included a lot of hard-to-classify instances. For many more features, it would be cognitively cumbersome for a human to explore all these features in different subspaces concurrently. The strategy that could be followed is to initially limit the space under examination using an extra preprocessing phase in the pipeline before utilizing FeatureEnVi for a detailed exploration of features. The overall scalability of the system could be improved in various ways, as previously described in~\autoref{sec:eval}.

\textbf{Other types of data.}
Regardless of the wide variety of application domains explored through the use cases and the case study, FeatureEnVi is not tested with other data types rather than structured tabular data consisting of numerical values~\cite{Shwartzziv2021Tabular}. 
One of our future intentions is to support further data types, but as our prototype tool is a proof of concept, the system's workflow and theoretical ideas are generalizable in this regard. 
Nonetheless, the features of each data set under investigation should be meaningful because we concentrate on human expertise and knowledge to resolve the contradictory situations between features and to prevent the generation of non-existent features for the sake of performance; in any other case, automatic feature engineering approaches might be preferable. For instance, if we consider image data transformed to numerical values, it is challenging to derive their specific meaning. Also, image data sets are typically combined with neural networks, which could not benefit as much from manual or semi-automatic feature engineering~\cite{Babaev2019E,ElKenawy2020Novel}.

\textbf{Targeted users.}
\hl{ML experts and practitioners are the main target groups that would benefit the most by using FeatureEnVi. We assume that they know at least the basics regarding the instances and features of their data, but they require further guidance to the feature engineering process. As seen in~\autoref{sec:eval}, the ML experts and the visualization researcher were able to understand and use FeatureEnVi within 1-hour interview sessions. An opportunity here is to develop another, more simplified version of our tool, specifically designed for novice users working with ML.}

\textbf{Completion time per activity.} 
The frontend of FeatureEnVi is implemented in JavaScript using Vue.js~\cite{vuejs}, D3.js~\cite{D3}, and Plotly.js~\cite{plotly}, and the backend is written in Python with Flask~\cite{Flask} and Scikit-learn~\cite{Pedregosa2011Scikit}. The use cases and experiments were performed on a MacBook Pro 2019 with a 2.6 GHz (6-Core) Intel Core i7 CPU, an AMD Radeon Pro 5300M 4 GB GPU, 16 GB of DDR4 RAM at 2667 Mhz, and running macOS Big Sur. By taking into consideration the specifications of the machine, we captured the total wall-clock time spent to complete each of the use cases and the case study (see~\autoref{time}, rows). The time measured is combined for the performance of the user actions and computational analyses as reported in every use case or case study (cf. Sections~\ref{sec:system} and~\ref{sec:case}). \autoref{time} columns map the time for each activity of feature engineering (i.e., selection, transformation, generation). Concretely, as the number of features analyzed increases, we observe an increase in the time required to compare the features and execute the user-defined commands. Interestingly, the feature transformation took longer in all cases, next was the generation, and finally the selection of subsets of features. These numbers can be rather subjective when different paths are tried out. In general, the efficiency of FeatureEnVi could be improved in numerous ways as already indicated before.

\begin{table}[tb]
\captionsetup{justification=centering, labelsep=newline}
\begin{threeparttable}
\caption{Analysis Completion Time for Each Activity of Feature Engineering for the Use Cases and the Case Study}
\label{time}
\setlength\tabcolsep{0pt} 
\begin{tabular*}{\columnwidth}{@{\extracolsep{\fill}} l ccc}
\toprule
     Data Set &
     \multicolumn{3}{c}{Feature Engineering} \\
\cmidrule{2-4}
     & Selection & Transformation & Generation \\
\midrule
     Red Wine Quality & 1:12 & 3:29 & 2:25 \\
     Vehicle Silhouettes & 2:58 & 6:14 & 5:24 \\
      QSAR Biodegradation & 5:30 & 33:24 & 13:35 \\
\bottomrule
\end{tabular*}
\smallskip
\scriptsize
\footnotesize{The wall-clock completion time is reported in the \emph{minute:second} format.}\\ \vspace{-4mm}
\end{threeparttable}
\end{table}

\section{Conclusions} \label{sec:con}%
  In this paper, we presented FeatureEnVi, a VA system with the aim to engineer features using stepwise feature selection and semi-automatic feature extraction approaches. Multiple coordinated views support users in selecting features, transforming them, and generating new features as part of an intensely iterative process. Exploring the impact of the features with several statistical measures and automatic feature selection techniques enables users to improve the predictive performance, reduce the need for computational resources, and decrease the time spent for training. Finally, our VA system is beneficial for feature engineering because it makes the complete process more transparent.
The effectiveness of FeatureEnVi was examined using real-world data sets that demonstrated improvements in performance with fewer available but optimally-tuned features.
Our system's workflow and visualizations obtained encouraging feedback from experts, who also helped us to identify the current limitations of FeatureEnVi. These limitations are considered as our future research directions.

\ifCLASSOPTIONcompsoc
  \section*{Acknowledgments}
  This work was partially supported through the ELLIIT environment for strategic research in Sweden. The authors also thank the anonymous reviewers for their constructive feedback.
\else
  \section*{Acknowledgment}
\fi





%



\bibliographystyle{IEEEtran}
\bibliography{references}

%
\vskip -2.5\baselineskip plus -1fil
\begin{IEEEbiography}[{\includegraphics[width=1in,height=1.25in,clip,keepaspectratio]{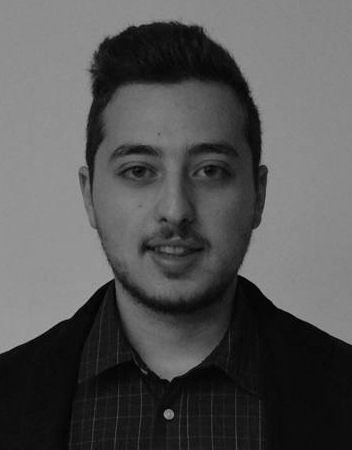}}]{Angelos Chatzimparmpas}
is PhD student within the ISOVIS research group and the Linnaeus University Centre for Data Intensive Sciences and Applications at the Department of Computer Science and Media Technology, Linnaeus University, Sweden. 
His main research interests include visual exploration of the inner parts and the quality of machine learning (ML) models with a specific focus on making complex ML models better understandable and explainable, as well as to provide reliable trust in the ML models and their results.
\end{IEEEbiography}
\vskip -2\baselineskip plus -1fil
\begin{IEEEbiography}[{\includegraphics[width=1in,height=1.25in,clip,keepaspectratio]{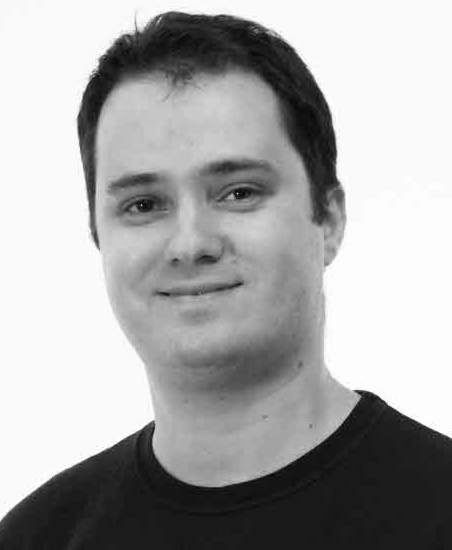}}]{Rafael M. Martins}
is Senior Lecturer at the Department of Computer Science and Media Technology at Linnaeus University, Sweden. His PhD research involved mainly the visual exploration of the quality of dimensionality reduction (DR) techniques, a topic he continues to investigate, in addition to other related research areas such as the interpretation of DR layouts and the application of DR techniques in different domains including software engineering and digital humanities.
\end{IEEEbiography}
\vskip -2\baselineskip plus -1fil
\begin{IEEEbiography}[{\includegraphics[width=1in,height=1.25in,clip,keepaspectratio]{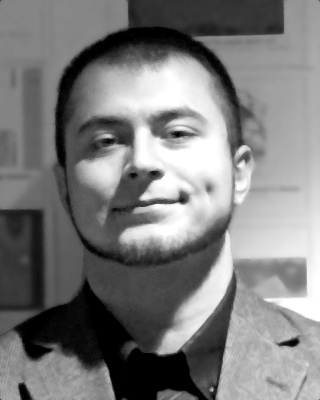}}]{Kostiantyn Kucher}
is Associate Senior Lecturer at the Department of Computer Science and Media Technology at Linnaeus University and part-time Visiting Assistant Lecturer at the Department of Science and Technology at Link{\"o}ping University, Sweden. His main research interests include visual text and network analytics, explainable AI, and domain applications of information visualization and visual analytics, especially in digital humanities and information science. 
\end{IEEEbiography}
\vskip -2\baselineskip plus -1fil
\begin{IEEEbiography}[{\includegraphics[width=1in,height=1.25in,clip,keepaspectratio]{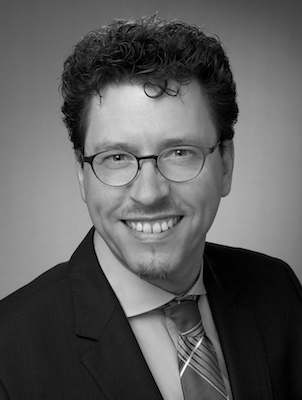}}]{Andreas Kerren}
is Professor of Information Visualization at Link{\"o}ping University (LiU) and Linnaeus University (LNU), Sweden. He holds the Chair of Information Visualization at LiU and is head of the ISOVIS research group at LNU. In addition, he is an ELLIIT professor supported by the Excellence Center at Link{\"o}ping--Lund in Information Technology and key researcher of the Linnaeus University Centre for Data Intensive Sciences and Applications. His main research interests include several areas of information visualization and visual analytics, especially visual network analytics, text visualization, and the use of visual analytics for explainable AI. 
\end{IEEEbiography}
\vskip -2\baselineskip plus -1fil
\vfill



\end{document}